\documentclass{article}

\usepackage{microtype}
\usepackage{graphicx}
\usepackage{subcaption}
\usepackage{booktabs} % for professional tables
\usepackage{xcolor}
\usepackage{tcolorbox}
\usepackage{array}
\usepackage{hyperref}
\usepackage{longtable}
\usepackage{xspace}
\usepackage{placeins}
\usepackage[tableposition=top]{caption}

\usepackage[accepted]{icml2024}

\usepackage{amsmath}
\usepackage{amssymb}
\usepackage{mathtools}
\usepackage{amsthm}

\usepackage{multirow}

\usepackage[capitalize,noabbrev]{cleveref}

\theoremstyle{plain}

\theoremstyle{definition}

\theoremstyle{remark}

\usepackage[textsize=tiny]{todonotes}

\icmltitlerunning{Superposition Prompting: Improving and Accelerating Retrieval-Augmented Generation}

\definecolor{titlecolor}{RGB}{0,0,0} % Change to your desired title color
\definecolor{myboxcolor}{RGB}{240,240,240}
\floatstyle{plain}
\floatname{myboxfloat}{Box}
\newtcolorbox{mybox}[2]{
  colback=myboxcolor,
  colframe=myboxcolor,
  fonttitle=\bfseries, % Bold title
  coltitle=titlecolor, % Font color of the title
}

\begin{document}

\twocolumn[
%\icmltitle{Schrödinger's Token: Improving Retrieval-Augmented Generation Performance with Superposition Prompting}
\icmltitle{Superposition Prompting: Improving and Accelerating Retrieval-\linebreak Augmented Generation}



%\icmlsetsymbol{equal}{}

\begin{icmlauthorlist}
\icmlauthor{Thomas Merth}{apple}
\icmlauthor{Qichen Fu}{apple}
\icmlauthor{Mohammad Rastegari*}{meta}
\icmlauthor{Mahyar Najibi}{apple}
\end{icmlauthorlist}

\icmlaffiliation{apple}{Apple, Cupertino, CA, USA}
\icmlaffiliation{meta}{Meta, Menlo xPark, CA, USA (*Work done while at Apple)}
%\icmlaffiliation{comp}{Company Name, Location, Country}
%\icmlaffiliation{sch}{School of ZZZ, Institute of WWW, Location, Country}

\icmlcorrespondingauthor{T. Merth}{tmerth@apple.com}
\icmlcorrespondingauthor{Q. Fu}{qfu22@apple.com}
% \icmlcorrespondingauthor{Mohammad Rastegari}{}
\icmlcorrespondingauthor{M. Rastegari}{mrastegari@meta.com}
\icmlcorrespondingauthor{M. Najibi}{najibi@apple.com}

\icmlkeywords{Machine Learning, ICML, LLM, Large Language Model, Transformer, Efficient ML, Dynamical Systems, Bayesian, In-Context Learning, Retrieval Augmented Generation, Question Answering}

\vskip 0.3in
]



\printAffiliationsAndNotice{}  % leave blank if no need to mention equal contribution

\begin{abstract}
Despite the successes of large language models (LLMs), they exhibit significant drawbacks, particularly when processing long contexts.
Their inference cost scales quadratically with respect to sequence length, making it expensive for deployment in some real-world text processing applications, such as retrieval-augmented generation (RAG).
Additionally, LLMs also exhibit the ``distraction phenomenon'', where irrelevant context in the prompt degrades output quality.
To address these drawbacks, we propose a novel RAG prompting methodology, \textit{superposition prompting}, which can be directly applied to pre-trained transformer-based LLMs \textit{without the need for fine-tuning}.
At a high level, superposition prompting allows the LLM to process input documents in parallel \textit{prompt paths}, discarding paths once they are deemed irrelevant.
We demonstrate the capability of our method to simultaneously enhance time efficiency across a variety of question-answering benchmarks using multiple pre-trained LLMs.
Furthermore, our technique significantly improves accuracy when the retrieved context is large relative the context the model was trained on.
For example, our approach facilitates a $93\times$ reduction in compute time while \textit{improving} accuracy by $43\%$ on the NaturalQuestions-Open dataset with the MPT-7B instruction-tuned model over naive RAG.
\end{abstract}

\newcommand{\pa}[1]{\ensuremath{\text{Pa}(#1)}}
\newcommand{\ch}[1]{\ensuremath{\text{Ch}(#1)}}

%%%%%%%%%%%%%%%%%%%%%%%%%%%%%%%%%%%%%%%%%%%%%%%%%%%%%%%%%%%%%%%%%%%%%%%%%%%%%%%
%%%%%%%%%%%%%%%%%%%%%%%%%%%%%%%%%%%%%%%%%%%%%%%%%%%%%%%%%%%%%%%%%%%%%%%%%%%%%%%
% MAIN PAPER
%%%%%%%%%%%%%%%%%%%%%%%%%%%%%%%%%%%%%%%%%%%%%%%%%%%%%%%%%%%%%%%%%%%%%%%%%%%%%%%
%%%%%%%%%%%%%%%%%%%%%%%%%%%%%%%%%%%%%%%%%%%%%%%%%%%%%%%%%%%%%%%%%%%%%%%%%%%%%%%
\setlength\fboxsep{1pt}

\section{Introduction}
\label{introduction}

\begin{figure}[ht]
    \centering
    \includegraphics[width=0.42 \textwidth]{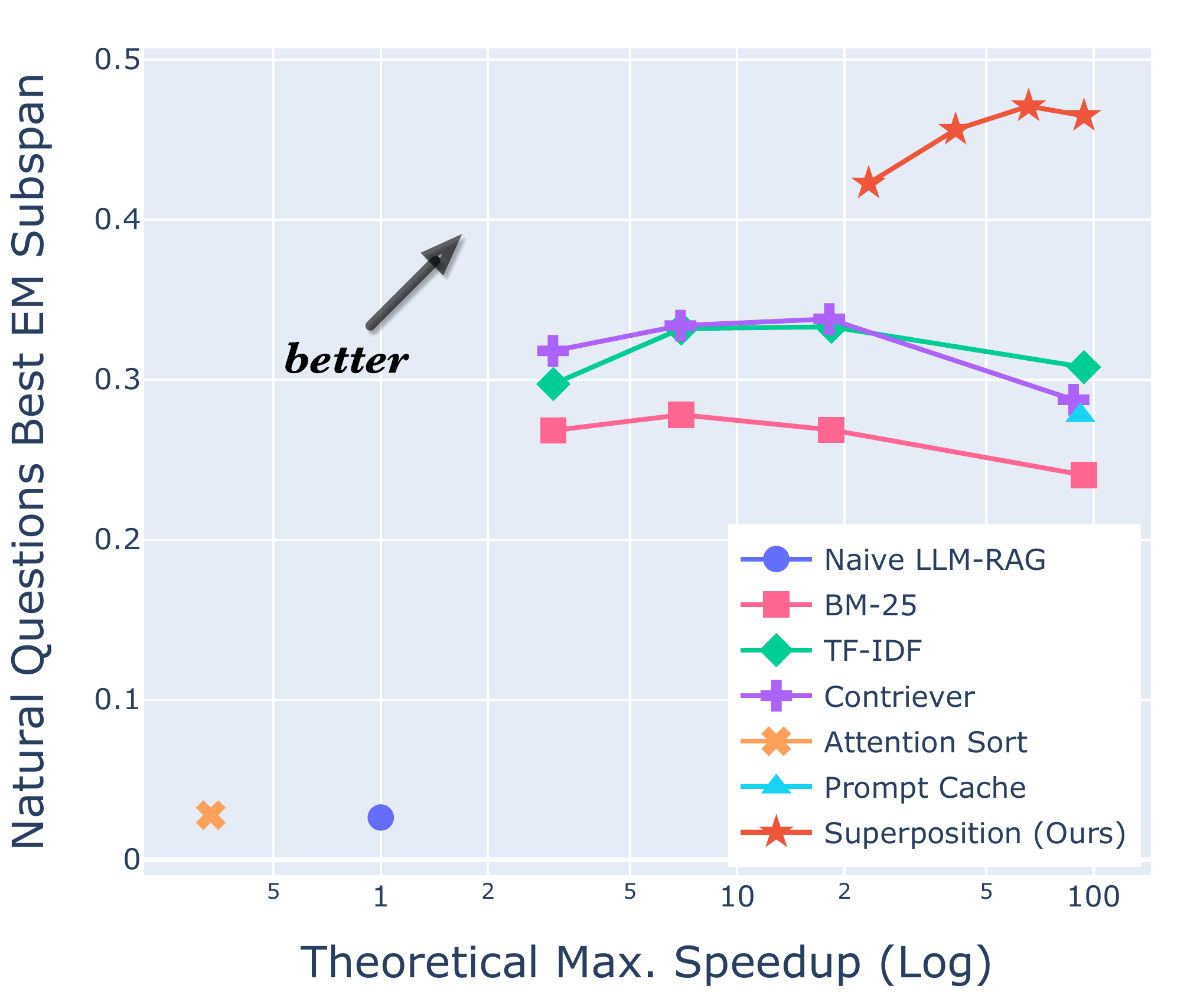}
    \caption{
        Theoretical Maximum Speedup vs. Accuracy (Best EM Subspan) on NaturalQuestions-Open using the \texttt{mpt-7b-instruct} model \cite{bloomz__Muennighoff2023CrosslingualGT}. Refer to \cref{section:experiment:nq} for experimental details.
        Plotted values are sourced from \cref{table:nq_baselines} and \cref{table:nq_topk_ablation}.
    }
    \label{figure:teaser_plot}
\end{figure}

\begin{figure*}[t]
    \centering
    \includegraphics[width=\textwidth]{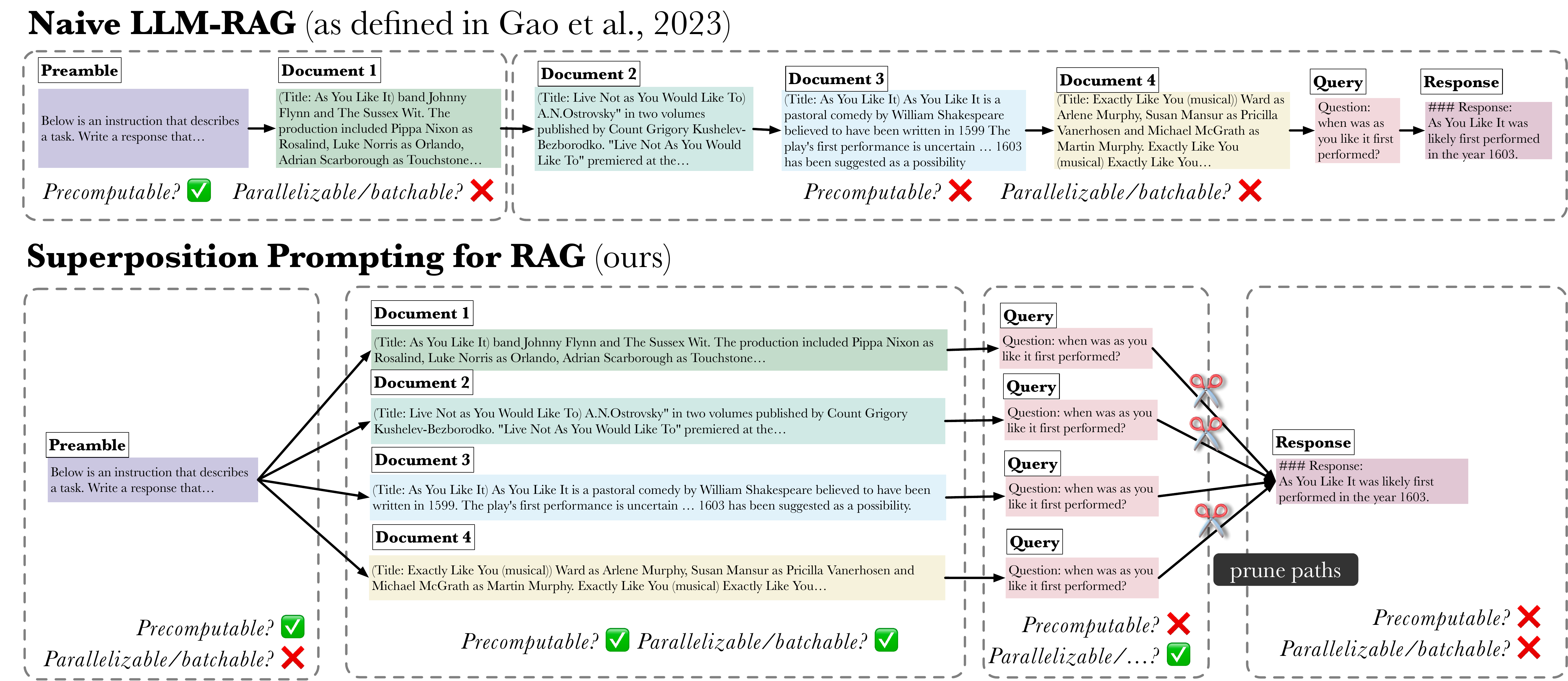}
    \caption{
        Comparison of superposition prompting vs. the ``classical'' (Naive LLM-RAG) prompting paradigm. Squares represents a token, and arrows depict attention dependencies.
        Whereas the classical approach is a ``linked-list'' style DAG, superposition prompting arranges token dependencies such that all documents are processed independently.
        Due to this dependency structure, we can easily leverage the LLM logits to prune irrelevant context, improving long context reasoning.
        The dependency structure also allows for faster prompt processing, due to the new opportunities for caching and parallelism of the KV cache and logit computations (each gray box represents, logically, a ``batch'' that is processed by the LLM, reusing upstream KV caches).
    }
    \label{figure:pdqtr_diagram}
\end{figure*}

Transformer-based autoregressive large language models (LLMs) have led to quantum leaps in text modeling performance over previous methods \cite{LLM_Survey__Zhao2023ASO}.
However, they have massive compute requirements, especially as the context length increases due to the quadratic compute cost of self-attention.
Many prior works have explored how to accelerate LLM inference \cite{LongContextSurvey__Huang2023AdvancingTA, EfficientServing__Miao2023TowardsEG}.
However, such optimizations often require significant architectural or parameter modifications to the pre-trained model, thus mandating expensive re-training or fine-tuning procedures.
In addition to causing undesirable compute requirements, long input contexts can also lead to hallucinations and/or divergent responses in model outputs \cite{liu2023lost, Distracted__Shi2023LargeLM}.

Retrieval-augmented generation (RAG) is one alluring application of transformer-based LLMs. In this setting, the LLM can ground its responses in auxiliary, relevant context.
Often, the retrieved documents contain long-form text, leading to the aforementioned downsides~\cite{RAGSurvey__Gao2023RetrievalAugmentedGF}.
To improve and accelerate RAG, we propose \textit{superposition prompting}
\footnote{
We drew inspiration from the ``path integral'' formulation of quantum mechanics \cite{PathIntegral}, where a particle's dynamics can be represented as a weighted sum over possible trajectories.
Analogously, the language dynamics of superposition prompting are modeled by a weighted sum of possible ``token trajectories''.
}.
Superposition prompting is demonstrated to simultaneously improve model accuracy and compute time efficiency for RAG-based question answering tasks \textit{without any additional training or fine-tuning}.
To highlight our results, we refer the reader to \cref{figure:teaser_plot}.

In this work, our contributions are as follows; (1) we propose a new generalized framework for prompting LLMs in RAG scenarios, (2) we demonstrate the benefit of our method on question-answering datasets, and (3) we provide extensive experimental evidence and ablation studies to give more confidence to our design decisions.
We also propose additional practical optimizations to accelerate inference by pruning, caching, and parallelizing the compute of \textit{prompt paths}. These optimizations are made possible due to the topological structure of our superposition prompts.

For reproducibility, our implementation can be found at \url{https://github.com/apple/ml-superposition-prompting}.

\section{Related Work}

\textbf{Retrieval Augmented Generation.} Retrieval-augmented generation (RAG) is a common application of LLMs to generate answers to questions based on a set of retrieved documents \cite{RAGOriginal__Lewis2020RetrievalAugmentedGF}.
Instead of simply prompting a language model with a query, RAG augments the prompt by injecting a set of retrieved documents into the prompt. If done correctly, these documents contain useful knowledge related to the query, which should elicit more accurate and reliable output from the model. Extensive work \cite{RAGOriginal__Lewis2020RetrievalAugmentedGF, guu2020realm, TokenRAG__Borgeaud2021ImprovingLM,RAGSurvey__Gao2023RetrievalAugmentedGF, SelfRAG__Asai2023SelfRAGLT} has shown RAG to be effective for many knowledge-intensive tasks \cite{petroni2020kilt}.
However, incorporating retrieved documents significantly extends the input sequence length and introduces additional computational overhead, raising efficiency concerns.
Addressing the challenges of long context processing and efficiency for RAG has become a key focus in recent research \cite{guu2020realm, beltagy2020longformer, PCW__Ratner2022ParallelCW}. 

\textbf{Efficient Long Context Processing.} There have been significant efforts to reduce the memory footprint and computational costs of transformers using techniques such as compression and KV-caching \cite{FlexGen__Sheng2023HighthroughputGI, AWQ__Lin2023AWQAW, SmoothQuant__Xiao2022SmoothQuantAA}.
More efficient versions of the transformer architecture have also been explored.
For example, Longformer \cite{beltagy2020longformer} introduced a drop-in replacement to the standard self-attention, which makes it scale linearly with sequence length.
Similarly, Reformer \cite{kitaev2020reformer} uses locality-sensitive hashing to reduce the complexity of attention and improve its efficiency for long sequences.
In parallel, the SparseTransformer \cite{child2019generating} focuses on the sparsity of the attention layers.
While the above innovation addresses the efficiency of long context processing, they often require non-trivial architecture changes and/or re-training.
This makes them impractical to use with existing, pre-trained LLMs \cite{touvron2023llama, zhang2022opt}.
Closer to our work are efficient methods which optimize KV caching and consider token importance \cite{HeavyHitter__Zhang2023H2OHO, Scissorhands__Liu2023ScissorhandsET}.
Other works (orthogonal to ours) investigate how to improve efficiency of LLM output generation \cite{SoT__Ning2023SkeletonofThoughtLL}.
The above methods differ from ours as they investigate acceleration for LLMs generally, whereas we aim to leverage the specifics of the RAG setting to achieve further improvements.

The closest to our work is the recently proposed Prompt Cache \cite{PromptCache__Gim2023PromptCM}. This method also leverages the modular structure of the RAG setting to perform local attention on the preamble and documents independently and cache the results. In contrast, our method retains attention dependencies \textit{between} segments in the form of a dependency graph. Also differentiating, we propose pruning and parallelization mechanisms not explored by \citealp{PromptCache__Gim2023PromptCM}.

\textbf{Prompt Engineering.}
Prompt engineering is the process of deliberately designing and tuning prompts before feeding them to language models to generate text \cite{PromptingSurvey__Liu2023PromptingFF}.
Prior exploration \cite{bubeck2023sparks} shows how careful prompt construction can greatly improve the model's responses.
Intriguingly, the recent work ``Lost in the Middle'' \cite{liu2023lost} has shown that solely the \textit{location} of the ``golden document'' (the document containing the answer) within a long context significantly affects the performance of language models.
Another theme of prompt engineering works has explored how to use graph-like structures while prompting LLMs.
Our proposed method might seem, at first glance, identical to other ``tree-based'' and ``graph-based'' prompting methods, such as Tree of Thoughts \cite{ToT__Yao2023TreeOT} and Graph of Thoughts \cite{GoT__Besta2023GraphOT}.
However, these methods are proposed in the context of multi-step reasoning, not RAG.
Different from the above, Parallel Context Windows \cite{PCW__Ratner2022ParallelCW}|along with other ``structured attention'' works \cite{StructuredAttn__Cai2023ScalingID, FID__Ye2023FiDICLAF}|aims to build dependencies between prompt text segments.
However, these works were generally applied to few-shot learning applications, not retrieval-augmented generation.
Our approach also differs from these structured attention papers in that we operate on generalized directed-acyclic graphs, as opposed to just the special case of trees.

%%%%%%%%%%%%%%%%%%%%%%%%%%%%%%%%%%%%%%%%%%%%%%%%%%%%%%%%%%%%%%%%%%%%%%%%%%%%%%%
% DEFS
%%%%%%%%%%%%%%%%%%%%%%%%%%%%%%%%%%%%%%%%%%%%%%%%%%%%%%%%%%%%%%%%%%%%%%%%%%%%%%%

\newcommand{\codeComment}[1]{$\triangleright$ \textit{#1}}

% Prompt graph objects
\newcommand{\tps}{\ensuremath{\mathcal{G}}}
\newcommand{\vertices}{\ensuremath{\mathcal{V}}}
\newcommand{\vertex}[1]{\ensuremath{\tilde{#1}}}
\newcommand{\vertexToTokenVal}[1]{\ensuremath{f(#1)}}
\newcommand{\tpEdges}{\ensuremath{\mathcal{E}}}

%% Named vertices
\newcommand{\preambleVertex}{\vertex{p}}
\newcommand{\docVertex}{\vertex{d}}
\newcommand{\queryVertex}{\vertex{q}}
\newcommand{\taskVertex}{\vertex{t}}

% Transformer objects
\newcommand{\vocabSize}{\ensuremath{n_v}}
\newcommand{\tokens}{\ensuremath{\mathcal{T}}}
\newcommand{\sequences}{\ensuremath{\mathcal{S}}}
\newcommand{\kvCaches}{\ensuremath{\mathcal{M}}}

% RAG-S
\newcommand{\nOfflineDocs}{\ensuremath{n_d}}
\newcommand{\docIndices}{\ensuremath{D}}

% Token sequences
\newcommand{\seqLen}[1]{\ensuremath{\|#1\|}}
\newcommand{\inputIds}[1]{\boldsymbol{#1}}
\newcommand{\docs}{\ensuremath{\boldsymbol{D}}}
\newcommand{\doc}{\inputIds{d}}
\newcommand{\preamble}{\inputIds{p}}
\newcommand{\query}{\inputIds{q}}
\newcommand{\queries}{\inputIds{Q}}
\newcommand{\task}{\inputIds{t}}
\newcommand{\response}{\inputIds{r}}
\newcommand{\new}{\inputIds{l}}

% Position Ids
\newcommand{\posIds}{\ensuremath{\mathcal{P}}}
\newcommand{\posId}[1]{\ensuremath{\check{#1}}}
\newcommand{\preamblePos}{\posId{p}}
\newcommand{\docPos}{\posId{d}}
\newcommand{\queryPos}{\posId{q}}
\newcommand{\taskPos}{\posId{t}}
\newcommand{\responsePos}{\posId{r}}
\newcommand{\newPos}{\posId{l}}

% KV caches
\newcommand{\kvCache}[1]{\ensuremath{\boxed{#1}}}
\newcommand{\emptyKv}{\kvCache{\emptyset}}
\newcommand{\preambleKv}{\kvCache{p}}
\newcommand{\docKv}{\kvCache{d}}
\newcommand{\docKvs}{\{\kvCache{d}\}_{j=1}^{\nOfflineDocs}}
\newcommand{\queryKv}{\kvCache{q}}
\newcommand{\taskKv}{\kvCache{t}}
\newcommand{\responseKv}{\kvCache{r}}
\newcommand{\newKv}{\kvCache{l}}

% LLM Call
\newcommand{\llmCall}{\ensuremath{\mathrm{LM}}}
\newcommand{\llmCallArgs}[3]{\ensuremath{\llmCall(#1, #2, #3)}}
\newcommand{\llmCallParallel}{\ensuremath{\mathrm{LMP}}}
\newcommand{\llmCallParallelArgs}[3]{\ensuremath{\llmCallParallel(#1, #2, #3)}}

% Logits.
\newcommand{\logits}{\ensuremath{\mathcal{L}}}
\newcommand{\logitVec}[1]{\ensuremath{#1}}
\newcommand{\preambleLogit}{\logitVec{\pi}}
\newcommand{\docLogit}{\logitVec{\delta}}
\newcommand{\queryLogit}{\logitVec{\phi}}
\newcommand{\taskLogit}{\logitVec{\tau}}
\newcommand{\responseLogit}{\logitVec{\rho}}
\newcommand{\newLogit}{\logitVec{\lambda}}

% XEnt
\newcommand{\crossEntName}{\ensuremath{H}}
\newcommand{\crossEntArgs}[2]{\ensuremath{\crossEntName(#1, #2)}}

% Tree
\newcommand{\treePromptName}{\textbf{Fork}}
\newcommand{\treePromptArgs}[2]{\treePromptName(#1, #2)}
\newcommand{\treePrompt}{\treePromptArgs{\preamble}{\docs}}
\newcommand{\treePromptVertices}{\ensuremath{V_<}}
\newcommand{\treePromptEdges}{\ensuremath{E_<}}

% Merge
\newcommand{\mergePromptName}{\textbf{Join}}
\newcommand{\mergePromptArgs}[2]{\ensuremath{\mergePromptName(#1, #2)}}
\newcommand{\mergePrompt}{\mergePromptArgs{\queries}{\task}}
\newcommand{\mergePromptVertices}{\ensuremath{V_>}}
\newcommand{\mergePromptEdges}{\ensuremath{E_>}}

% TreeMerge
\newcommand{\treeMergePromptName}{ForkJoin}
\newcommand{\treeMergePromptArgs}[4]{\treeMergePromptName(#1, #2, #3, #4)}
\newcommand{\treeMergePrompt}{\treeMergePromptArgs{\preamble}{\docs}{\query}{\task}}
\newcommand{\treeMergePromptVertices}{\ensuremath{V_{<>}}}
\newcommand{\treeMergePromptEdges}{\ensuremath{E_{<>}}}

\newcommand{\computePosName}{\textbf{Equilibrium}}
\newcommand{\computePosArgs}[1]{\computePosName(#1)}

\begin{figure*}[ht]
\centering
\subfloat[Baseline]
{
    \includegraphics[width=73pt]{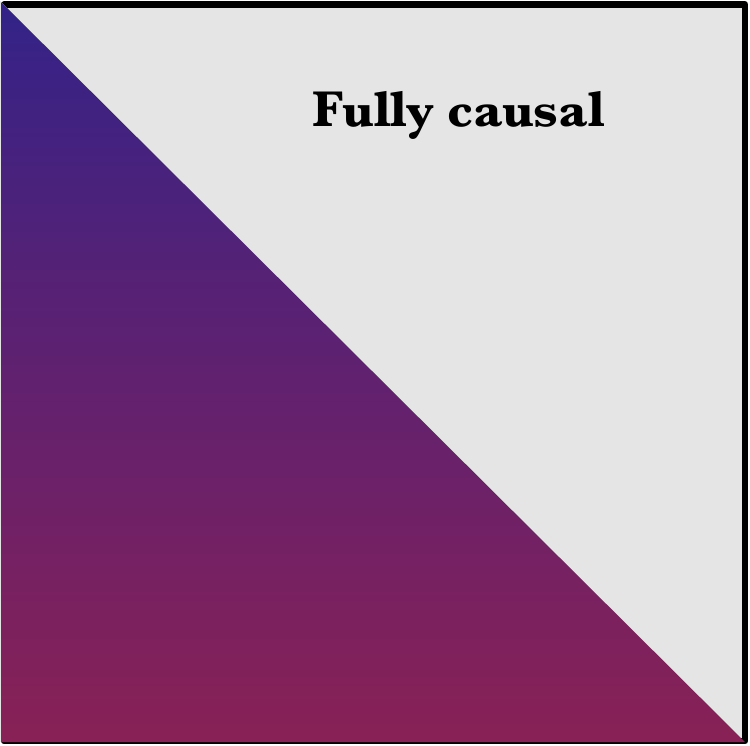}
    \label{figure:visual_overview:baselin}
}
\subfloat[Superposition prompting]
{
    \includegraphics[width=73pt]{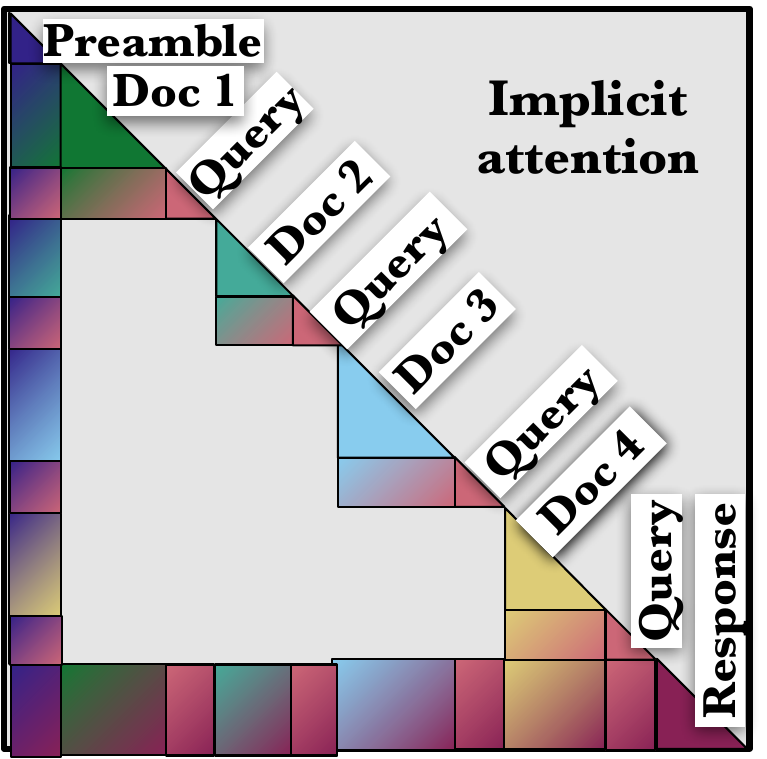}
    \label{figure:visual_overview:vanilla}
}
\subfloat[w/ Path Pruning]
{
    \includegraphics[width=73pt]{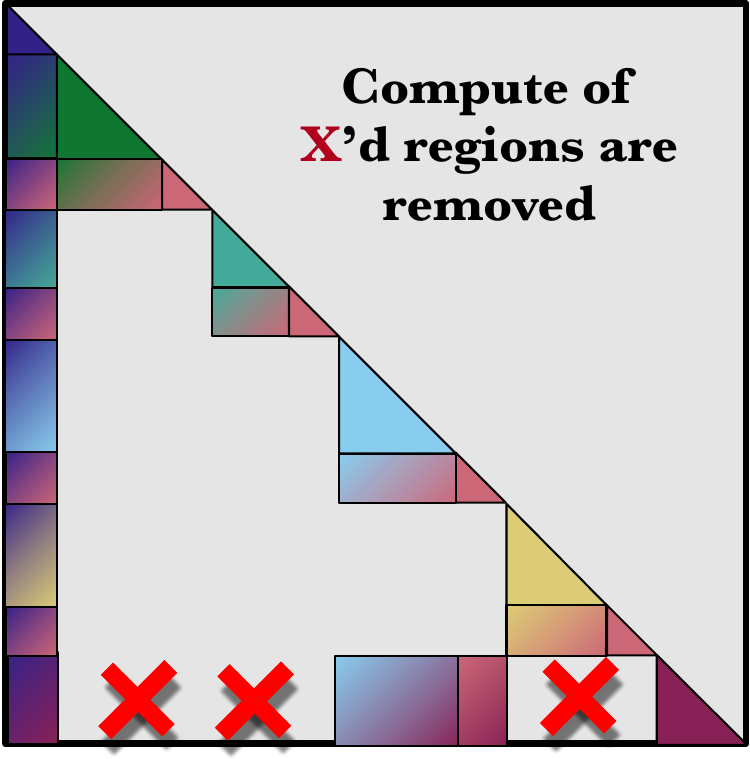}
    \label{figure:visual_overview:pruning}
}
\subfloat[w/ Path Caching]
{
    \includegraphics[width=73pt]{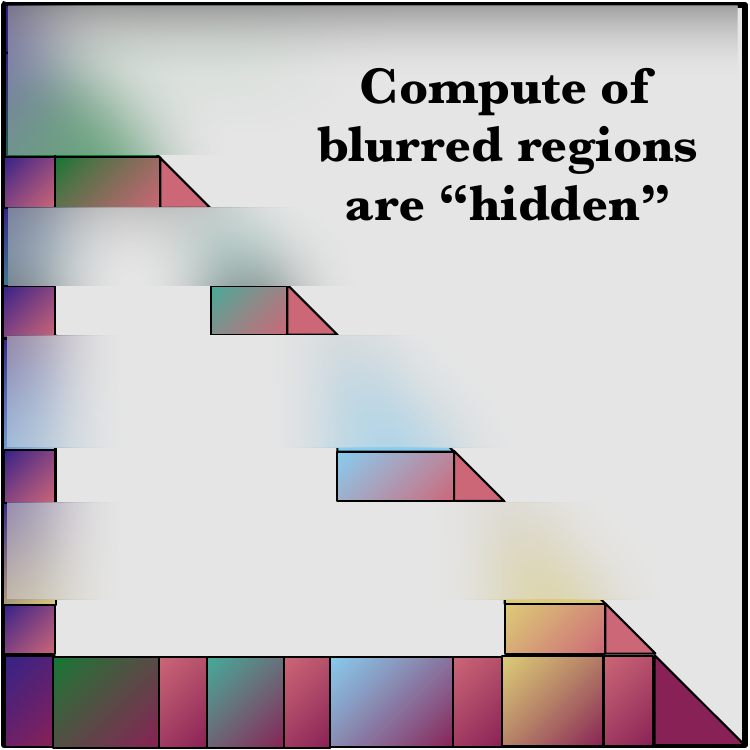}
    \label{figure:visual_overview:caching}
}
\subfloat[w/ Path Parallelization]
{
    \includegraphics[width=73pt]{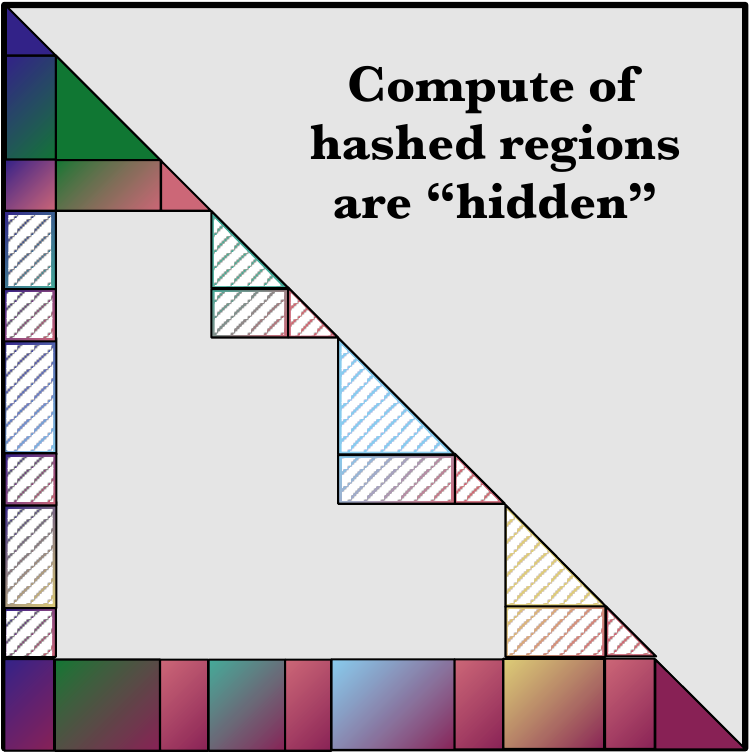}
    \label{figure:visual_overview:parallel}
}
\subfloat[w/ All]
{
    \includegraphics[width=73pt]{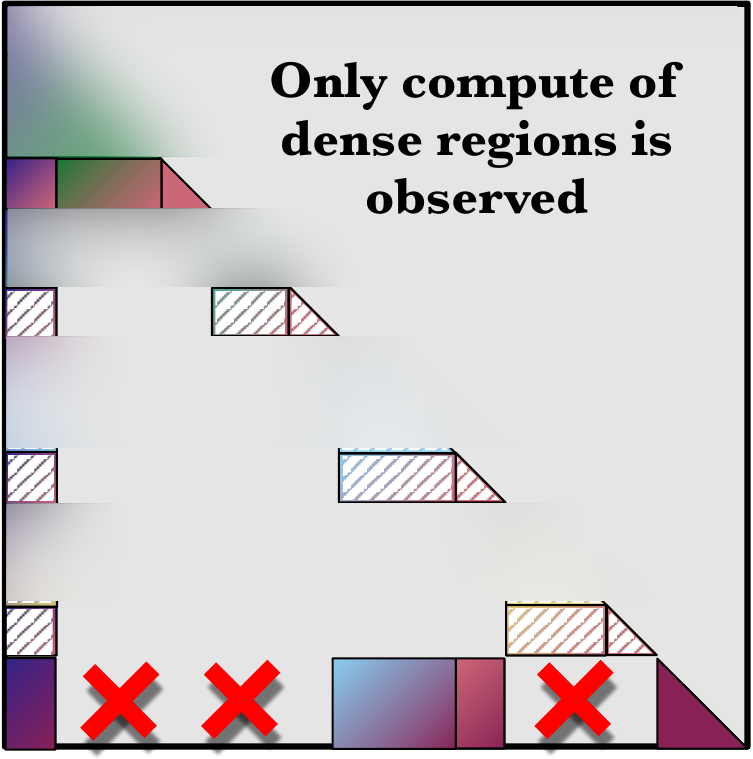}
    \label{figure:visual_overview:all}
}
\caption{
    Implicit attention dependencies that must be computed during ``online serving'' (the colors in (b)-(f) correspond to the token segment colors in \cref{figure:pdqtr_diagram}).
    Note how the various optimizations reduce the computational burden required at online serving-time by pruning, precomputing, and parallelizing the work.
    It is worth re-emphasizing that in practice, inference is \textit{not} sparse attention on one large sequence, but rather \textit{dense attention} with many different shorter token segments.
}
\label{figure:visual_overview}
\end{figure*}

\section{Proposed Method}
\label{superposition_prompting}

The retrieval-augmented generation task is comprised of distinct text segments|the preamble (\textit{a.k.a.} system prompt), a (static) corpus of documents, and a (dynamic) user-supplied query.
Instead of concatenating these text segments in textual space, we group them into separate ``batches'' (the gray boxes in \cref{figure:pdqtr_diagram}), which are passed as calls to the LLM (re-using the KV caches from upstream token segments).
With a query as input, superposition prompting processes all choices of documents paired with the query independently (conditioned on the preamble)|in \cref{figure:pdqtr_diagram}, this can be seen as the branching structure.
Once the query batch is processed, we then employ \textit{path pruning} (\cref{section:configurations:pruning}) to discard entire attention dependencies based on an importance metric (the scissors in \cref{figure:pdqtr_diagram}).
Both of these optimizations improve inference efficiency and enable the model to discard distracting documents unrelated to the query.

Enabled by the added structure of our superposition prompting approach, we then propose techniques to further accelerate the inference.
First, the high-level notion of token sharing across prompts allows us to employ \textit{prompt path caching} (\cref{section:configurations:caching}).
Finally, we describe a \textit{prompt path parallelization} (\cref{section:configurations:parallelization}) strategy that leverages independence across segments.

\subsection{Retrieval Augmented Generation}
\label{section:method:rag}

We stylize token sequences as bolded vectors and use $\oplus$ to denote concatenation along the sequence dimension.
Supposing there are $\nOfflineDocs$ (pre-tokenized) offline documents available for retrieval, we define the set of document token sequences $\{\doc_1, \ldots, \doc_{\nOfflineDocs}\}$.
We denote the user query as $\query$, and our custom preamble sequence as $\preamble$.
The goal is to return some response $\response$ which answers the query, all while minimizing the latency between the arrival of the query and the serving of the response as observed by the client.
The obvious baseline solution (which we refer to as ``Naive LLM-RAG'') is where one simply concatenates the input sequences as $\inputIds{x} = \preamble \oplus \doc_1 \oplus \cdots \oplus \doc_{\nOfflineDocs} \oplus \query$, then autoregressively generates $\response$ using $\inputIds{x}$ as the prompt. However, as shown in \cref{sec:experimental_results}, our approach massively outperforms such a baseline both in terms of quality and performance.

\subsection{Superposition Prompting}
\label{section:method:overview}

We now detail superposition prompting, a new paradigm for prompting language models.
In superposition prompting, prompts are \textit{not} represented as a simple sequence of tokens as they are with ``classical'' prompting methods (\textit{e.g.} Naive LLM-RAG).
Rather, superpositioned prompts are directed acyclic graphs (DAGs) where nodes are token sequences, and edges codify attention dependencies.
Plainly put, a particular token sequence, $\inputIds{v}$, attends to the tokens in another token sequence, $\inputIds{u}$, if and only if there is a path from $\inputIds{u}$ to $\inputIds{v}$ in the DAG.
In this sense, superposition prompting is a generalization of ``classical'' prompting (since a ``classical'' prompt is the linked-list special case).
Please refer to \cref{alg:serving} for an algorithmic formalization.

\subsubsection{The \treeMergePromptName\xspace Prompt Path Topology}
In order to leverage superposition prompting for RAG, we must construct a graph topology out of our text segments.
We propose the \textit{\treeMergePromptName} graph structure, depicted in \cref{figure:pdqtr_diagram}.
Note that each $\query_i$ sequence is a duplicate of the original $\query$ (\cref{section:configurations:pruning} will justify this decision).
Although this duplication ends up increasing the number of tokens processed, our ablation analysis in \cref{appendix:algorithm:bayesian} demonstrates the superiority of this approach in terms of accuracy.
Furthermore, \cref{section:configurations:parallelization} describes how the cost of this duplication can be entirely hidden from the user.
The \treeMergePromptName\xspace topology (implicitly) results in a pseudo-``local attention'' structure (\cref{figure:visual_overview}).
We emphasize that this resulting attention pattern is a construct for visualization only|in reality, all calls to the LLM use fully dense attention, although on relatively smaller context lengths.
Concretely, each of the dashed boxes in \cref{figure:pdqtr_diagram} is a separate LLM call.

\subsubsection{Token Position Assignment}
\label{section:position_assignment}

With classical prompting, tokens are (by default) spaced equally, with a distance of $1$.
However, with superposition prompting, positioning tokens is not trivial, since paths (of potentially heterogeneous length) run parallel to each other.
Thus, we seek to answer the question, ``how do we assign meaningful positions to tokens in superposition prompts?''

One simple approach could be to truncate token sequences to a common length to enforce a shared positioning.
However, truncating may result in loss of input signal if the trimmed tokens contained valuable information for the query.
Another approach could be to \textit{left-align} (or right-pad) sequences to a common length~\cite{PromptCache__Gim2023PromptCM}.
While this \textit{left aligned} padding approach is simple, it yields discontinuities in the prompt sequence position assignments (see \cref{appendix:stats} for quantification).
With the ALiBi encoding scheme \cite{ALiBi__DBLP:journals/corr/abs-2108-12409}, it can be easily shown that discontinuities unfairly assign attention penalties to the tokens in shorter documents, since the tokens will be concentrated at earlier token positions (and thus larger distances from the current token).\footnote{
An analogous bias would exist if using a \textit{right aligned} strategy, except shorter documents would be unfairly assigned attention boosts over longer documents.
}
Thus, we are motivated to propose a positional assignment strategy that does not result in discontinuities.

\begin{figure}[ht]
\centering
\subfloat[Left Aligned]
{
    \includegraphics[width=110pt]{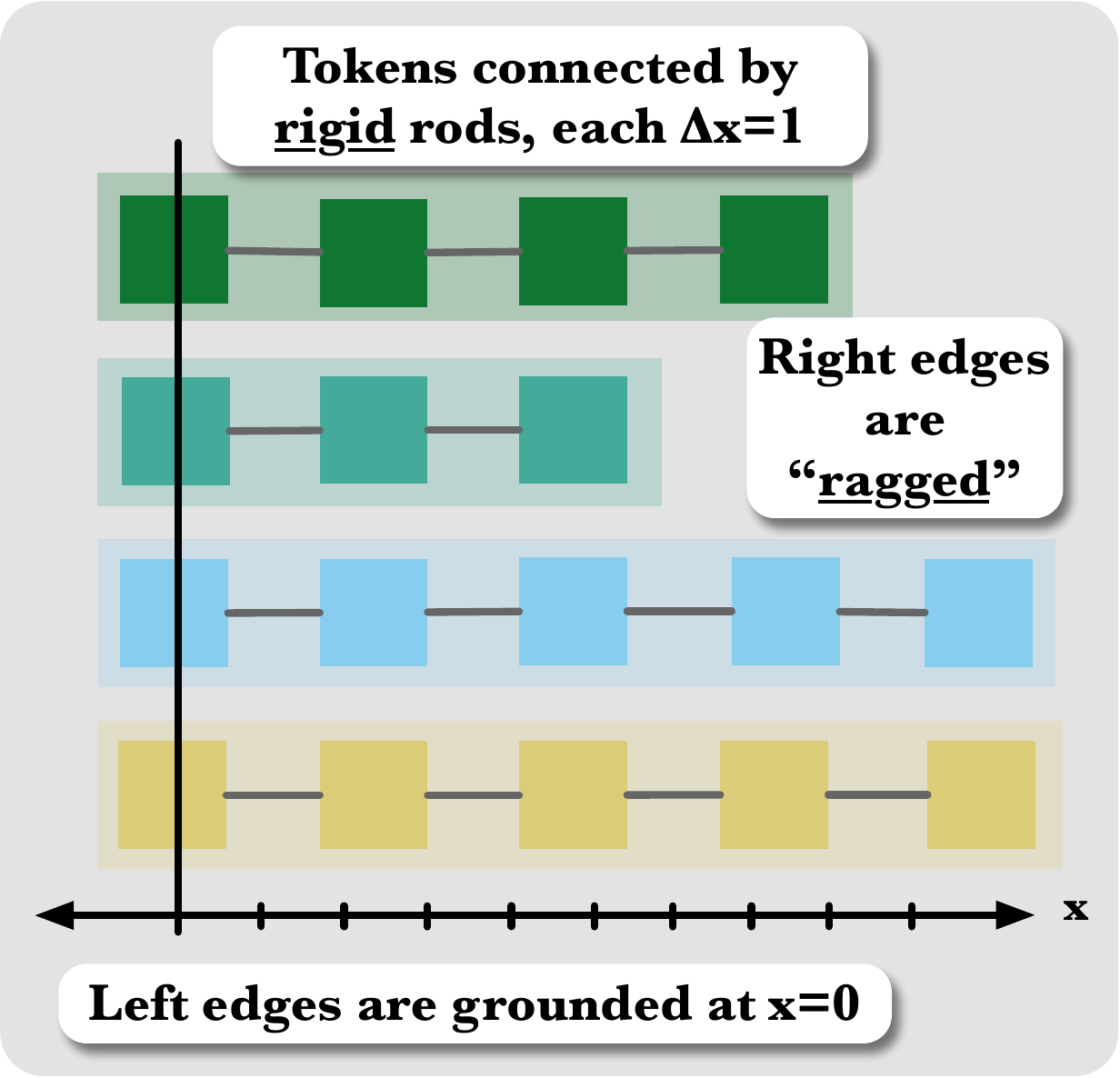}
    \label{figure:pos:left_align}
}
\subfloat[Equilibrium (Ours)]
{
    \includegraphics[width=118pt]{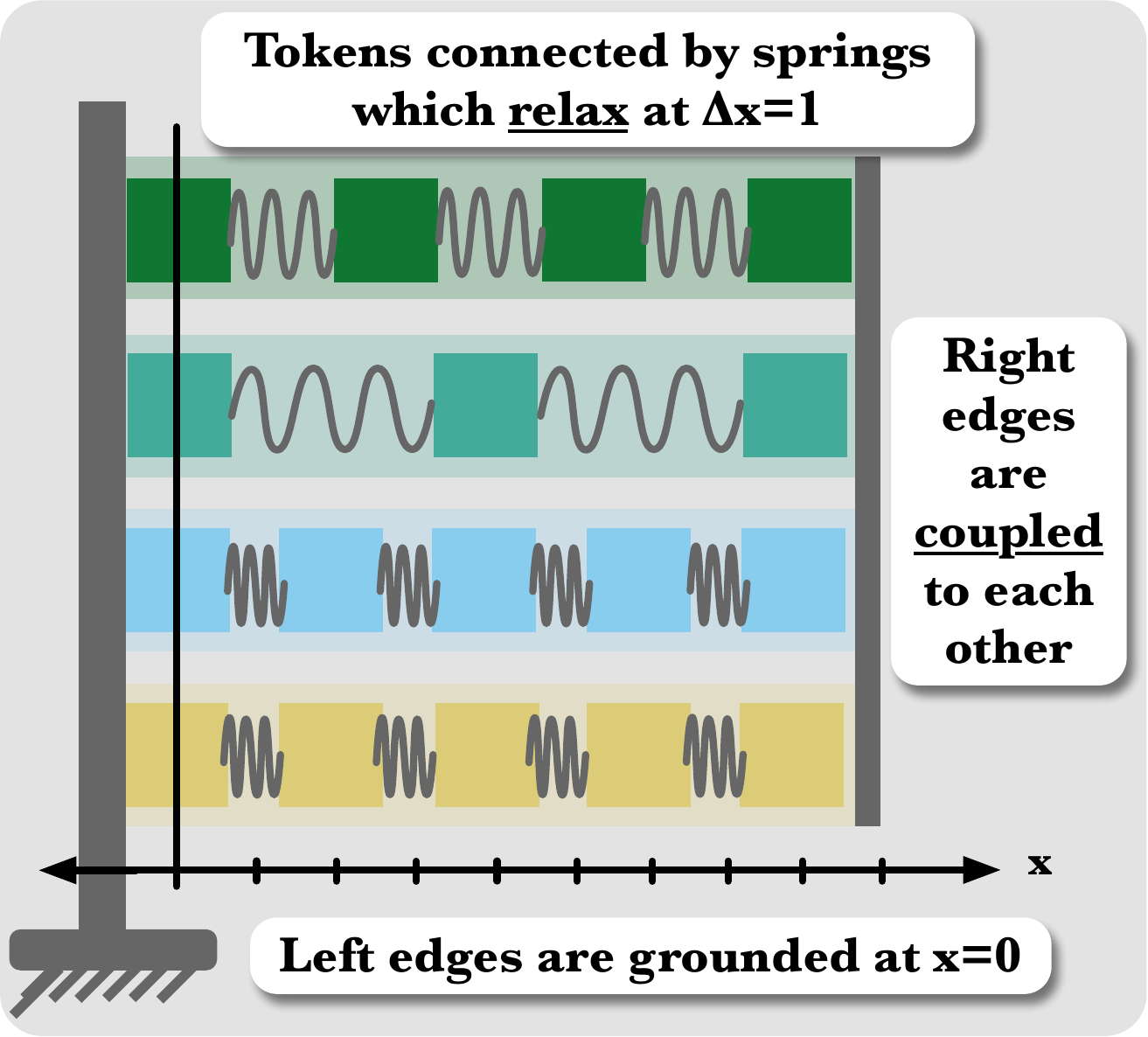}
    \label{figure:pos:equilibrium}
}
\caption{Visual intuition for our proposed \textit{equilibrium position assignment} vs. left aligned (see \cref{section:position_assignment}).
}
\label{figure:positioning}
\end{figure}

We propose \textit{path equilibrium positioning} as one simple, reasonable strategy.
With path equilibrium positioning, we linearly space overlapping paths to fit the harmonic mean, $S(\docs)$, of their collective lengths (for a set of overlapping paths $\docs$)
\begin{equation}
    S(\docs) = \frac{\nOfflineDocs}{\sum_{\doc \in \docs} \frac{1}{\seqLen{\doc}}}
    \label{equation:distance}
\end{equation}
Intuitively, the resulting token positions matches the equilibrium state of coupled masses connected by springs (\cref{figure:positioning}).

Note that the path equilibrium positioning strategy results in real-valued positions.
This is a departure from common usage of token position assignments, where integer-valued positions are predominant.\footnote{
While this is trivial for the ALiBi positional encoding, it is non-trivial for the Rotary Position Embedding \cite{RoPE__Su2021RoFormerET} scheme.
To handle this case, we linearly interpolate the \textit{argument} of the sinusoidal functions.
}
We note that the choice of position assignment scheme has no effect on inference efficiency, but can impact model output quality.
In \cref{appendix:ablations:nq_pos_ablation}, we validate the superiority of path equilibrium positioning.

\subsubsection{Path Pruning}
\label{section:configurations:pruning}

Further exploiting the topological structure we've imposed on the prompt, we propose \textit{path pruning} as a mechanism to discard documents it sees as irrelevant to the query.
As demonstrated in our experiments, this can benefit both efficiency and accuracy for LLM-based RAG.

In order to prune paths, we must compute a saliency score for each path.
Inspired by SGPT \cite{SGPT__Muennighoff2022SGPTGS}, we apply Bayes rule to the output of the language modeling head to compute a saliency or entailment score.
At a high level, we leverage Bayes' theorem to compute the posterior distribution $$P(\doc_i \mid \query_i, \preamble) \propto P(\query \mid \doc_i, \preamble) P(\doc_i \mid \preamble)$$ as a saliency metric of document $\doc_i$'s relevancy to the query.\footnote{
This resembles the ``principle of least action,'' which determines the optimal path weighting in the path integral formulation of quantum mechanics.
}
In our experiments, we decide which path indices to prune by greedily selecting the top-$k$ of this categorical distribution (we perform ablations with respect to the choice of $k$ in \cref{section:experiments:musique} and \cref{appendix:ablations:topk}).

To ``physically'' apply the pruning, we can simply discard the KV caches corresponding to the documents and queries along those paths.
Conveniently, all remaining KV caches can be simply concatenated together for use in autoregressive generation of the response.

We provide ablations against other reasonable saliency metrics in \cref{appendix:ablations:path_pruning}.
A visual representation of the effect of path pruning on the (implicit) attention patterns can be also seen in \cref{figure:visual_overview:pruning}.

\subsection{Lossless Runtime Optimizations}
\label{section:configurations}

\subsubsection{Path Caching}
\label{section:configurations:caching}
Assuming auxiliary memory storage is available, we can accelerate inference of superposition prompts by doing work before any query has arrived.
This \textit{path caching} technique generalizes the ideas put forth in PagedAttention \cite{PagedAttention__Kwon2023EfficientMM}, where we cache the KV embeddings along all path prefixes (not just the ``root node'', as PagedAttention does). Importantly, our approach also differs from PromptCache \cite{PromptCache__Gim2023PromptCM}.
While their cached ``prompt modules'' only attend locally to themselves, our path prefix KV caches attend locally to themselves \textit{as well as} to all their ancestors in the graph.
Please refer to \cref{alg:preprocessing} in appendix for formalization.

We now describe our path caching mechanism.
Concretely, the preamble KV cache and document KVs are not conditioned on the query, and thus can be precomputed during a ``preprocessing'' stage.
Then, during the ``online serving'' stage, we retrieve the preamble and document KVs from storage instead of the original input token sequences.
\cref{figure:visual_overview:caching} shows how, during the online serving stage, much of the attention dependencies have already been computed.
Note that the memory requirement for employing path caching is a scalar multiple, $c_\textrm{model}$, of the raw tokenized sequence length. Here, $c_\textrm{model}$ is a fixed scalar that depends on the various aspects of the models, such as number of layers, and embedding dimension (\textit{e.g.} $c_\texttt{bloom-7b1} = 492 \mathrm{\ KB}$).

\subsubsection{Path Parallelization}
\label{section:configurations:parallelization}

Since the superpositioned paths of \textit{\treeMergePromptName}\xspace are independent of each other (by construction), the corresponding KV caches and logits of the query segments can be computed in parallel.
While this does not reduce the ``CPU time,'' it importantly reduces the \textit{wall-clock time} experienced by the user.
The parallelization across the duplicated queries can be accomplished either by (1) concatenating sequences along the batch dimension before inference\footnote{
In general, prompt path lengths will vary, thus requiring padding to a common length. However, a length binning strategy may alleviate most overhead in practice.
} (2) delegating model calls across a distributed cluster of compute nodes (\textit{e.g.} GPUs), or (3) a combination of batching and distributed inference.
The most effective strategy will depend on the specifics of the cluster configuration (\textit{e.g.} relative network bandwidth vs. available compute per node).

\section{Experimental Results}
\label{sec:experimental_results}

We perform experiments on three families of large language models, namely OpenELM~\cite{Mehta2024OpenELMAE}, BLOOMZ~\cite{bloomz__Muennighoff2023CrosslingualGT}, and MPT~\cite{MosaicML2023Introducing}. 
To quantify the effectiveness of superposition prompting when paired with models of different scales, we use various model sizes from these families.
For OpenELM, we use the \texttt{3B-Instruct} configuration.
For BLOOMZ, we instantiate 3B parameter (\texttt{bloomz-3b}) and 7.1B parameter (\texttt{bloomz-7b1}) models.
Finally, for MPT, we use the available instruct fine-tuned 7B parameter model (\texttt{mpt-7b-instruct}).
This set of models covers different architectures, positional encoding schemes, sizes, and pretraining recipes.
We remind the reader we use the publicly released pretrained checkpoints, without employing any additional training, fine-tuning, or task adaptation.

For our experiments, we are primarily interested in the compute time vs. accuracy tradeoff.\footnote{
In our timing and speedup analysis, we follow previous works \cite{PromptCache__Gim2023PromptCM} and do not consider the data retrieval portion of the RAG pipeline, which would require too many assumptions.
}
We use the \texttt{fvcore} \cite{fvcore} package to compute theoretical floating point operation (FLOP) counts for various inference settings. We evaluate the compute cost of each method in units of \textit{compute cycles}|similar to FLOPs, but accounting for parallelism.
In practice, to achieve the speedups, extra resources (auxiliary memory and/or auxiliary compute for parallelization) will be required.
However, as stated, the goal of this exploration is \textit{acceleration}, not necessarily FLOPs reduction.
We refer the reader to \cref{appendix:ablations:opt} for detailed breakdowns of the theoretical speedup gains enabled by each of our proposed optimizations.

\subsection{Results}

We leverage the publicly available NaturalQuestions-Open \cite{liu2023lost} and MuSiQue \cite{trivedi2022musique} datasets.
We do not perform any manual prompt tuning or prompt engineering for any method or baseline, and use the same prompts across all experiments (per dataset) to control for discrepancies that could arise with varying prompt wording.
For reproducibility, we present the exact prompt wording used for each dataset in \cref{appendix:prompt_examples}.
We use greedy autoregressive decoding in all experiments, and randomize the order of documents to prevent any systematic bias possible due to location of the ``gold documents'' (à la \citealp{liu2023lost}).

\subsubsection{NaturalQuestions-Open}

\begin{table*}[thb]
\caption{Retrieval augmented generation accuracy for various models and methods on the NaturalQuestions-Open dataset. For baselines with hyperparameters---namely the top-$k$ parameter for BM-25, TF-IDF, and Contriever---we present their highest accuracy configuration (see \cref{appendix:ablations:topk} for all configurations). We emphasize the superiority of superposition prompting over the considered baselines along the axes of both accuracy and speedup.}
\label{table:nq_baselines}
\vskip 0.15in
\begin{center}
\begin{small}
\begin{sc} 
\begin{tabular}{llcccc}
\toprule
 &  & Compute Cycles & Theor. Speedup & Accuracy \\
Model & Approach &  &  &  \\
\midrule
\multirow{7}{*}{OpenELM-3B-In.} & Naive LLM-RAG & 1.03e+13 & 1.0 & 0.001 \\
 & BM-25 & 1.07e+11 & 96.9 & 0.166 \\
 & TF-IDF & 1.07e+11 & 96.9 & 0.215 \\
 & Contriever & 5.62e+11 & 18.4 & 0.191 \\
 & Attention Sort & 3.09e+13 & 0.3 & 0.000 \\
 & Prompt Cache & 1.25e+11 & 82.4 & 0.000 \\
 & Superposition (Ours) & \textbf{1.07e+11} & \textbf{96.8} & \textbf{0.241} \\
\cline{1-5}
\multirow{7}{*}{bloomz-3b} & Naive LLM-RAG & 9.75e+12 & 1.0 & 0.005 \\
 & BM-25 & 1.35e+12 & 7.2 & 0.138 \\
 & TF-IDF & 1.35e+12 & 7.2 & 0.188 \\
 & Contriever & 1.37e+12 & 7.1 & 0.168 \\
 & Attention Sort & 2.93e+13 & 0.3 & 0.004 \\
 & Prompt Cache & 1.29e+11 & 75.4 & 0.113 \\
 & Superposition (Ours) & \textbf{9.92e+10} & \textbf{98.3} & \textbf{0.223} \\
\cline{1-5}
\multirow{7}{*}{bloomz-7b1} & Naive LLM-RAG & 2.22e+13 & 1.0 & 0.022 \\
 & BM-25 & 7.35e+12 & 3.0 & 0.150 \\
 & TF-IDF & 3.21e+12 & 6.9 & 0.203 \\
 & Contriever & 3.23e+12 & 6.9 & 0.194 \\
 & Attention Sort & 6.67e+13 & 0.3 & 0.022 \\
 & Prompt Cache & 2.86e+11 & 77.7 & 0.136 \\
 & Superposition (Ours) & \textbf{2.38e+11} & \textbf{93.5} & \textbf{0.253} \\
\cline{1-5}
\multirow{7}{*}{mpt-7b-instruct} & Naive LLM-RAG & 2.16e+13 & 1.0 & 0.026 \\
 & BM-25 & 3.11e+12 & 7.0 & 0.278 \\
 & TF-IDF & 1.18e+12 & 18.4 & 0.333 \\
 & Contriever & 1.20e+12 & 18.1 & 0.338 \\
 & Attention Sort & 6.49e+13 & 0.3 & 0.028 \\
 & Prompt Cache & 2.36e+11 & 91.8 & 0.278 \\
 & Superposition (Ours) & \textbf{2.31e+11} & \textbf{93.7} & \textbf{0.465} \\
\cline{1-5}
\bottomrule
\end{tabular}
\end{sc}
\end{small}
\end{center}
\end{table*}

\label{section:experiment:nq}
NaturalQuestions-Open~\cite{liu2023lost} is an open domain question answering benchmark that is derived from Natural Questions \cite{kwiatkowski-etal-2019-natural}.
It contains the historical queries issued to the Google search engine, coupled with answers using the contents of English Wikipedia.
We follow the same experimental setup as \citealp{liu2023lost}, including the same preprocessing and evaluation methodology for the 20 document setting (reporting Best EM Subspan, or ``Accuracy'' for short).

We present speedup vs. accuracy comparisons in \cref{table:nq_baselines}. For the TF-IDF baseline, we use TF-IDF (from SciPy package \citealp{2020SciPy-NMeth}) to select the top-$k$ documents conditioned on the query, then perform ``naive LLM-RAG'' (as described in \cref{section:method:rag}).
Our BM-25 baseline is equivalent, except we use \citealp{rank_bm25} for the top-$k$ document selection.
We also have an equivalent baseline where we use Contriever \cite{Contriever__Izacard2021UnsupervisedDI} to select the top-$k$ documents.\footnote{
For a more generous representation of the BM-25, TF-IDF, and Contriever baselines, we compute the speedup metrics assuming document KV caching (although to our knowledge, this has not been previously proposed in literature).
Note that caching is not possible with Naive LLM-RAG or Attention Sort since the order of documents is variable, and in general, documents attend to other documents (thus also parallelization is not possible).
}
We compare against the recently proposed Attention Sort method, using their method exactly as described in \citealp{AttentionSort__Peysakhovich2023AttentionSC}.
Finally, we compare against Prompt Cache \cite{PromptCache__Gim2023PromptCM}.
Note that Naive LLM-RAG, Prompt Cache, and Attention Sort always attend to all documents.

In addition to \cref{table:nq_baselines}, we present various architectural ablation studies in \cref{section:ablations} and \cref{appendix:ablations} to justify our design decisions.

\subsubsection{MuSiQue}
\label{section:experiments:musique}

MuSiQue~\cite{trivedi2022musique} is a multi-hop reasoning dataset consisting of question answer pairs collected with the goal of making disconnected reasoning harder and consequently adding to the difficulty of the previously introduced multi-hop question answering datasets. We validate our approach on the dev split of MuSiQue-Ans (reporting Answer EM and F1).

A slight modification is made to superposition prompting to handle the multi-hop reasoning setup of MuSiQue.
Specifically, we iteratively apply superposition pruning to build a chain of $t \times k$ documents\footnote{
Note that only the first $k$ documents chosen are cacheable. Subsequent documents are not cacheable since their KV caches depend on preceding documents (which are dynamically chosen during query serving).
}, where $t$ and $k$ are hyperparameters.
At each time step $\{1, \ldots t\}$, we create a superposition with all remaining documents, prune to keep the top $k$, prepend those (cached) documents to the running prefix, then repeat.
A visual depiction of this \textit{iterative superposition} is presented in \cref{fig:iterative_superposition}.
We hypothesize that iterative superposition can improve performance since we equip the LLM to iteratively solve the multi-hop reasoning challenge.

For our baselines, we compare against Attention Sort, Prompt Cache, and Naive LLM-RAG (all of which always attend to all documents).
Our results are summarized in \cref{table:musique_baselines}.

\begingroup
\renewcommand{\arraystretch}{1.1}
\begin{table*}[thb]
\caption{Retrieval augmented generation accuracy for various models on the MuSiQue dataset. For superposition prompting, $t$ denotes the number of iterations of iterative superposition (described in \cref{section:experiments:musique}), and $k$ denotes the top-$k$ selected (i.e. not pruned) at each step (see \cref{section:configurations:pruning}).}
\label{table:musique_baselines}
\vskip 0.15in
\begin{center}
\begin{small}
\begin{sc}\begin{tabular}{llcccc}
\toprule
 &  & Compute Cycles & Theor. Speedup & F1 & EM \\
Model & Approach &  &  &  &  \\
\midrule
\multirow{6}{*}{OpenELM-3B-In.} & Naive LLM-RAG & 8.81e+12 & 1.0 & 0.006 & 0.000 \\
 & Attention Sort & 2.64e+13 & 0.3 & 0.009 & 0.001 \\
 & Prompt Cache & 1.61e+11 & 54.7 & 0.000 & 0.000 \\
 & Superposition (X=1, K=4) (Ours) & \textbf{1.42e+11} & \textbf{62.1} & 0.028 & 0.000 \\
 & Superposition (X=2, K=4) (Ours) & 5.97e+11 & 14.8 & 0.039 & 0.006 \\
 & Superposition (X=4, K=1) (Ours) & 1.99e+12 & 4.4 & \textbf{0.059} & \textbf{0.019} \\
\cline{1-6}
\multirow{6}{*}{bloomz-3b} & Naive LLM-RAG & 8.31e+12 & 1.0 & 0.060 & 0.030 \\
 & Attention Sort & 2.49e+13 & 0.3 & 0.055 & 0.026 \\
 & Prompt Cache & 1.64e+11 & 50.5 & 0.136 & 0.081 \\
 & Superposition (X=1, K=4) (Ours) & \textbf{1.33e+11} & \textbf{62.4} & 0.173 & 0.100 \\
 & Superposition (X=2, K=4) (Ours) & 5.56e+11 & 14.9 & \textbf{0.187} & \textbf{0.117} \\
 & Superposition (X=4, K=1) (Ours) & 1.87e+12 & 4.4 & 0.187 & 0.115 \\
\cline{1-6}
\multirow{6}{*}{bloomz-7b1} & Naive LLM-RAG & 1.91e+13 & 1.0 & 0.062 & 0.033 \\
 & Attention Sort & 5.72e+13 & 0.3 & 0.058 & 0.031 \\
 & Prompt Cache & 3.67e+11 & 52.0 & 0.161 & 0.108 \\
 & Superposition (X=1, K=4) (Ours) & \textbf{3.17e+11} & \textbf{60.2} & 0.159 & 0.090 \\
 & Superposition (X=2, K=4) (Ours) & 1.33e+12 & 14.3 & 0.171 & 0.106 \\
 & Superposition (X=4, K=1) (Ours) & 4.44e+12 & 4.3 & \textbf{0.182} & \textbf{0.115} \\
\cline{1-6}
\multirow{6}{*}{mpt-7b-instruct} & Naive LLM-RAG & 1.86e+13 & 1.0 & 0.064 & 0.008 \\
 & Attention Sort & 5.57e+13 & 0.3 & 0.062 & 0.009 \\
 & Prompt Cache & 3.09e+11 & 60.1 & 0.088 & 0.018 \\
 & Superposition (X=1, K=4) (Ours) & \textbf{3.04e+11} & \textbf{61.1} & 0.111 & 0.029 \\
 & Superposition (X=2, K=4) (Ours) & 1.29e+12 & 14.4 & 0.111 & 0.031 \\
 & Superposition (X=4, K=1) (Ours) & 4.26e+12 & 4.4 & \textbf{0.120} & \textbf{0.040} \\
\cline{1-6}
\bottomrule
\end{tabular}
\end{sc}
\end{small}
\end{center}
\end{table*}

\subsection{Analysis}

\subsubsection{Superposition Prompting Can Improve Time Efficiency}

Results on the NaturalQuestions-Open dataset \cref{table:nq_baselines} shows that superposition prompting is the leading efficient method by an order of magnitude.
These gains are mainly due to the path parallelism, and path pruning mechanisms. \cref{table:nq_opt_ablation_small} presents a breakdown of the contribution of each of these mechanisms to the speedup.
For instance, for \texttt{mpt-7b-instruct} (on NaturalQuestions-Open), caching alone yields a $10.2\times$ speedup, whereas parallelism alone yields a $14.8\times$ speedup.
These optimizations combined with pruning yield a $93.7\times$ speedup overall.

With MuSiQue, we see observe lower overall speedups for the highest performing superposition prompting settings (\cref{table:musique_baselines}).
This is due to the employment of \textit{iterative superposition} (\cref{section:experiments:musique}), which limits caching opportunities to the selection of the first $k$ documents.

\subsubsection{Superposition Prompting Can Improve Accuracy}

In \cref{table:nq_baselines} we clearly see that superposition prompting is the dominant method in terms of accuracy on NaturalQuestions-Open, seeing improvements of $12$--$43\%$ over the naive solution, and up to $15\%$ improvements over the next best competitor.
With MuSiQue (\cref{table:musique_baselines}), we note that superposition prompting yields the highest accuracy for each model.

One explanation for the accuracy improvement is how superposition prompting reduces sequence lengths as perceived by the transformer.
Recent studies have investigated the apparent lack of ``length extrapolation'' abilities of transformer-based LLMs \cite{ALiBi__DBLP:journals/corr/abs-2108-12409, RandomPosEnc__Ruoss2023RandomizedPE, PosEnc__Kazemnejad2023TheIO}.
One convenient property of superposition prompting is that|from the perspective of the transformer|the maximum sequence length observed is the \textit{longest path through the graph}.\footnote{
This means the effective (perceived) sequence length is $\mathcal{O}(1)$ instead of $\mathcal{O}(\nOfflineDocs)$, where $\nOfflineDocs$ is the number of offline documents.
}
For example, with NaturalQuestions-Open, superposition prompting decreases the maximum path (and thus the sequence length) from an average of 2923 tokens to 206 tokens.
In this sense, superposition prompting for RAG can enable \textit{non}-long-context transformers to perform well on long sequences.
This property could allow model developers to significantly reduce pretraining costs (since training special-purpose ``long-context'' LLMs leads to increased costs \cite{ALiBi__DBLP:journals/corr/abs-2108-12409}).

Another explanation for the accuracy improvement is the LLM ``distraction'' phenomenon.
The previous works of \citealp{liu2023lost, RETRO__Borgeaud2021ImprovingLM, Distracted__Shi2023LargeLM} present arguments for how LLMs can be sensitive to noisy or irrelevant context.
With the inclusion of the path pruning mechanism, we equip the model with a structured way to filter out the ``noise'' (\textit{i.e. irrelevant documents}).

\subsubsection{Sensitivity to ALiBi vs. RoPE}

For reasons outlined in \cref{section:position_assignment}, superposition prompting is very naturally suited for transformers which accept continuously-valued token position assignments (\textit{i.e.} they exhibit the \textit{position interpolation} property as defined by \citealp{PositionInterp__Chen2023ExtendingCW}).
While the ALiBi positional encoding scheme has been shown to posses this property, it has been suggested that fine-tuning may be required to equip Rotary Position Embedding (RoPE)-based models with this property.

Our experiments validate that our proposed equilibrium positional assignment mechanism is compatible even with a \textit{non}-fine-tuned RoPE-based model (\textit{i.e.} the OpenELM family).
We leave it to future studies to measure the extent to which fine-tuning may improve accuracy (if at all).

We note that \texttt{OpenELM-3B-Instruct} has significantly lower accuracy for many baselines, such as AttentionSort, Naive LLM-RAG and even Prompt Cache.
We hypothesize that this is due to the lack of length extrapolation capabilities of RoPE, which would become more pronounced for those baselines.

\begingroup
\renewcommand{\arraystretch}{1}
\begin{table}[tbh]
\caption{Varying the position assignment function used with superposition prompting on the NaturalQuestions-Open dataset.}
\label{table:nq_pos_ablation}
\begin{center}
\begin{small}
\begin{sc}\begin{tabular}{llc}
\toprule
 &  & Accuracy \\
Model & Approach &  \\
\midrule
\multirow{2}{*}{OpenELM-3B-In.} & Left Aligned & 0.224 \\
 & Equilibrium (Ours) & \textbf{0.241} \\
\cline{1-3}
\multirow{2}{*}{bloomz-3b} & Left Aligned & 0.208 \\
 & Equilibrium (Ours) & \textbf{0.223} \\
\cline{1-3}
\multirow{2}{*}{bloomz-7b1} & Left Aligned & 0.245 \\
 & Equilibrium (Ours) & \textbf{0.253} \\
\cline{1-3}
\multirow{2}{*}{mpt-7b-instruct} & Left Aligned & 0.348 \\
 & Equilibrium (Ours) & \textbf{0.465} \\
\cline{1-3}
\bottomrule
\end{tabular}
\end{sc}
\end{small}
\end{center}
\end{table}

\section{Ablations}
\label{section:ablations}

\subsection{Position Assignment Ablation}
\label{appendix:ablations:nq_pos_ablation}
In \cref{table:nq_pos_ablation}, we investigate the effect of the position assignment strategy during superposition prompting.
We compare our proposed equilibrium path positioning to the left aligned strategy described in \cref{section:position_assignment}.
Our findings validate our hypothesis outlined in \cref{alg:equilibrium}, where we speculated that left alignment would result in worse performance (due to long sequence attention bias).

\subsection{Path Saliency Metric Ablation}
\label{appendix:ablations:path_pruning}

In \cref{table:nq_attention_ablation}, we ablate our choice of ``path saliency'' metric. We compare against two other baselines|\textit{attention} and \textit{none}.
With \textit{none}, we simply do not prune.
The \textit{attention} baseline consists of using the attention scores for each document (average across tokens, layers and attention heads) as the path score.
We highlight that our Bayesian path saliency significantly outperforms attention-based scoring, as well as the control baseline.

\begin{table}[H]
\caption{Retrieval augmented generation accuracy for various path saliency metrics on the NaturalQuestions-Open dataset.}
\label{table:nq_attention_ablation}
\vskip 0.15in
\begin{center}
\begin{small}
\begin{sc}\begin{tabular}{llccc}
\toprule
 &  & Accuracy \\
Model & Selection Metric &  \\
\midrule
\multirow{3}{*}{OpenELM-3B-In.} & None & 0.005 \\
 & Attention & 0.163 \\
 & Bayesian (Ours) & \textbf{0.241} \\
\cline{1-3}
\multirow{3}{*}{bloomz-3b} & None & 0.110 \\
 & Attention & 0.100 \\
 & Bayesian (Ours) & \textbf{0.223} \\
\cline{1-3}
\multirow{3}{*}{bloomz-7b1} & None & 0.142 \\
 & Attention & 0.127 \\
 & Bayesian (Ours) & \textbf{0.253} \\
\cline{1-3}
\multirow{3}{*}{mpt-7b-instruct} & None & 0.224 \\
 & Attention & 0.218 \\
 & Bayesian (Ours) & \textbf{0.465} \\
\cline{1-3}
\bottomrule
\end{tabular}
\end{sc}
\end{small}
\end{center}
\end{table}

\subsection{Superposition Factor Ablation}
\label{appendix:ablation:superposition_factor}

We introduce the hyperparameter \textit{superposition factor} as a parameter to interpolate between a fully superimposed and fully ``classical'' prompt.
Larger superposition factors correspond to ``more superimposed'' prompts, whereas smaller superposition factors correspond to ``less superimposed'' prompts (achieved by combining adjacent documents before creating prompt paths).

Formally, we define $m$ as the number of documents considered for a retrieval-augmented generation query (for instance, this is $m = 20$ for common settings of NaturalQuestions-Open \cite{liu2023lost} and MuSiQue \cite{trivedi2022musique}).
By setting a superposition factor $\gamma \in [1, m]$, we compute the ``effective documents per path'' as $\lfloor \frac{m}{\gamma} \rceil$.
Importantly, note that when $\gamma = 1$, we've reduced to the ``classical'' (Naive LLM-RAG) case. We perform an ablation by sweeping this superposition factor parameter and present results in \cref{fig:nq_superposition_quotient_plot}.
A visual representation is presented in \cref{fig:superposition_quotient}.

The curves generally show improvements along both axes as we increase the superposition quotient.
Interestingly, the maximal accuracy may not be fully superimposed, suggesting that this value should be tuned for the given application.

\begin{figure}[H]
    \centering
    \includegraphics[width=220pt]{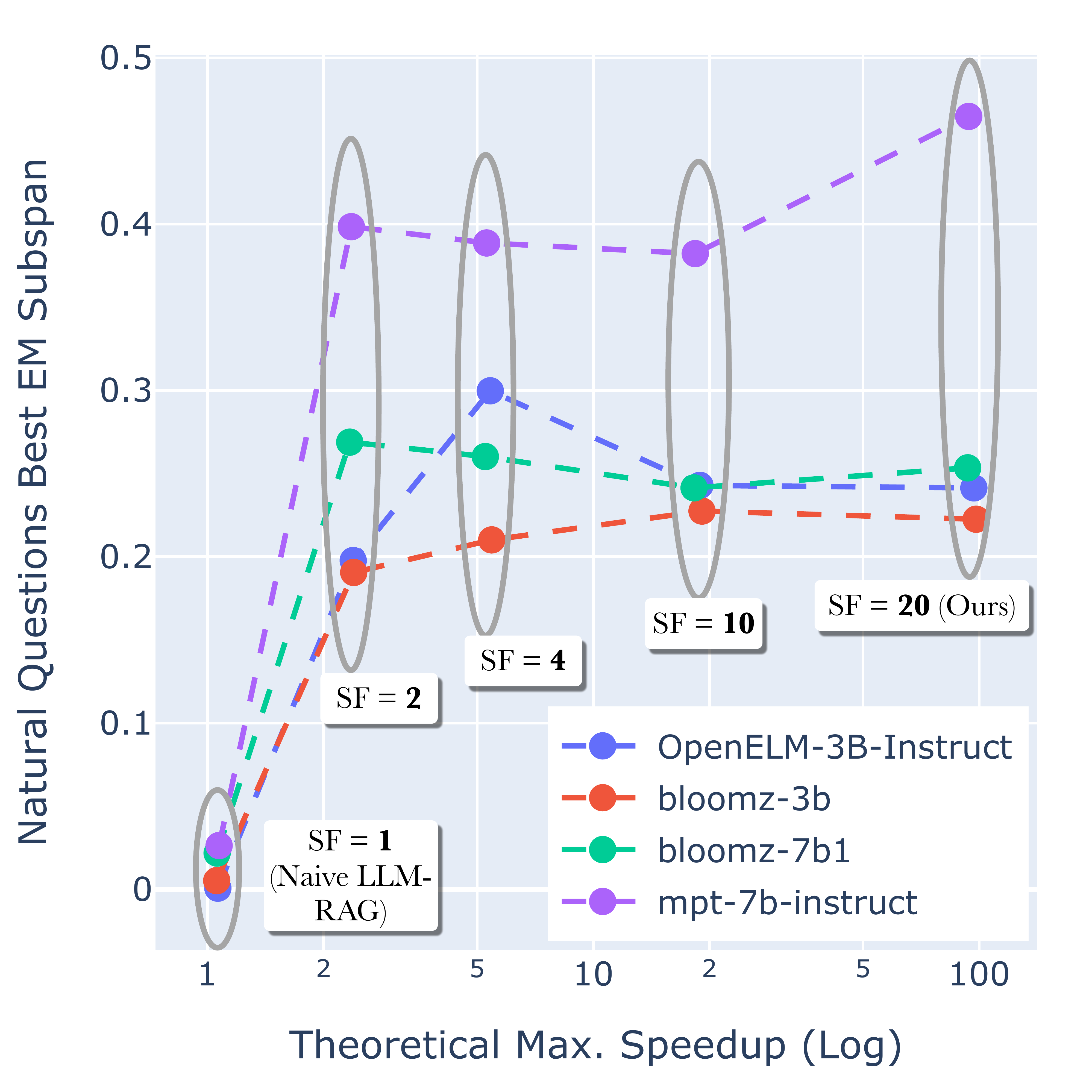}
    \caption{
        Sweeping values of superposition factor (SF) on the NaturalQuestions-Open dataset with a variety of models.
    }
    \label{fig:nq_superposition_quotient_plot}
\end{figure}

\section{Conclusion and Discussion}
In this work, we introduced a novel framework to accelerate and improve retrieval-augmented generation with LLMs.
We verified the generalization of our method across various models and datasets and performed extensive ablations.

Our method, \textit{superposition prompting}, was shown to improve long sequence modeling accuracy in single and multi-hop question answering tasks, all while reducing user-observed response latency.
Conveniently, this optimization was shown to be effective without any fine-tuning or additional training to the base model.
We defer to future work to explore how (if at all) fine-tuning could \textit{further} improve superposition prompting.
We also highlight that future work should investigate how to generalize these ideas outside of the RAG setting.

\section*{Impact Statement}
This paper presents work whose goal is to advance the field of Machine Learning. Specifically, we are proposing a machine learning technique that can be used for improving the quality of generative language modeling while reducing the computational overheads often associated with their deployment. There are many potential consequences of this work, as for the field as a whole, none which we feel must be specifically highlighted here.

\section*{Acknowledgements}
We would like to extend a thanks to Sachin Mehta, Maxwell Horton, Enrico Fini, and Arsalan Farooq for discussion and feedback on the paper.

%%%%%%%%%%%%%%%%%%%%%%%%%%%%%%%%%%%%%%%%%%%%%%%%%%%%%%%%%%%%%%%%%%%%%%%%%%%%%%%
% No-cite references.
%%%%%%%%%%%%%%%%%%%%%%%%%%%%%%%%%%%%%%%%%%%%%%%%%%%%%%%%%%%%%%%%%%%%%%%%%%%%%%%

\bibliography{main}
\bibliographystyle{icml2024}

%%%%%%%%%%%%%%%%%%%%%%%%%%%%%%%%%%%%%%%%%%%%%%%%%%%%%%%%%%%%%%%%%%%%%%%%%%%%%%%
%%%%%%%%%%%%%%%%%%%%%%%%%%%%%%%%%%%%%%%%%%%%%%%%%%%%%%%%%%%%%%%%%%%%%%%%%%%%%%%
% APPENDIX
%%%%%%%%%%%%%%%%%%%%%%%%%%%%%%%%%%%%%%%%%%%%%%%%%%%%%%%%%%%%%%%%%%%%%%%%%%%%%%%
%%%%%%%%%%%%%%%%%%%%%%%%%%%%%%%%%%%%%%%%%%%%%%%%%%%%%%%%%%%%%%%%%%%%%%%%%%%%%%%
\newpage
\appendix
\onecolumn

\section{Ablations (cont.)}
\label{appendix:ablations}

\subsection{Top-$k$ Ablation}
\label{appendix:ablations:topk}

Here, we sweep values for top-$k$ for our method, where $k$ are the number of documents retained for generating the answer (full table results are provided in \cref{table:nq_topk_ablation}).
For completeness, we also ablate this hyperparameter for all ``ranking-based'' baselines, namely BM-25, TF-IDF, and Contriever (described in \cref{section:experiment:nq}).
Note that for superposition prompting, this top-$k$ value corresponds the number of paths retained after path pruning.
We see that accuracy tends to peak around $k=2$ to $k=4$, although it comes at a cost to speedup versus $k=1$.
Interestingly we do see that performance decreases steadily for increasing $k > 4$.
This seems to coincide with the arguments put forth in \cite{Distracted__Shi2023LargeLM} and \citealp{ALiBi__DBLP:journals/corr/abs-2108-12409} and \citealp{liu2023lost} of ``increased distraction'' as the context length increases.

\begin{table}[H]
\caption{Retrieval augmented generation accuracy and speedups for various models on the NaturalQuestions-Open dataset, sweeping top-$k$.
Note that ``Comp.'' is used as an abbreviation for ``Compute'', and ``Sp.'' is used as an abbreviation for ``Speedup''.
}
\label{table:nq_topk_ablation}
\begin{center}
\begin{small}
\begin{sc}
\setlength\extrarowheight{-2pt}
\begin{tabular}{llccccc}

\toprule
 &  &  & Comp. Cycles & Theor. Sp. & CUDA Sp. & Accuracy \\
Model & Approach & Top-k &  &  &  &  \\
\midrule
\multirow{16}{*}{bloomz-3b} & \multirow{4}{*}{BM-25} & 8 & 3.13e+12 & 3.1 & 2.40 & 0.129 \\
 &  & 4 & 1.35e+12 & 7.2 & 3.83 & 0.138 \\
 &  & 2 & 5.09e+11 & 19.1 & 5.53 & 0.106 \\
 &  & 1 & 9.92e+10 & 98.2 & 6.51 & 0.092 \\
\cline{2-7}
 & \multirow{4}{*}{Contriever} & 8 & 3.15e+12 & 3.1 & 2.26 & 0.162 \\
 &  & 4 & 1.37e+12 & 7.1 & 3.51 & 0.168 \\
 &  & 2 & 5.25e+11 & 18.6 & 4.87 & 0.153 \\
 &  & 1 & 1.15e+11 & 84.5 & 5.61 & 0.114 \\
\cline{2-7}
 & \multirow{4}{*}{TF-IDF} & 8 & 3.13e+12 & 3.1 & 2.40 & 0.164 \\
 &  & 4 & 1.35e+12 & 7.2 & 3.81 & 0.188 \\
 &  & 2 & 5.09e+11 & 19.1 & 5.48 & 0.187 \\
 &  & 1 & 9.92e+10 & 98.2 & 6.46 & 0.155 \\
\cline{2-7}
 & \multirow{4}{*}{Superposition (Ours)} & 8 & 4.06e+11 & 24.0 & 5.80 & 0.244 \\
 &  & 4 & 2.30e+11 & 42.4 & 5.92 & \textbf{0.264} \\
 &  & 2 & 1.43e+11 & 68.3 & 5.96 & 0.252 \\
 &  & 1 & 9.92e+10 & 98.2 & 5.64 & 0.223 \\
\cline{1-7} \cline{2-7}
\multirow{16}{*}{bloomz-7b1} & \multirow{4}{*}{BM-25} & 8 & 7.35e+12 & 3.0 & 2.33 & 0.150 \\
 &  & 4 & 3.21e+12 & 6.9 & 3.99 & 0.150 \\
 &  & 2 & 1.22e+12 & 18.2 & 5.40 & 0.128 \\
 &  & 1 & 2.38e+11 & 93.4 & 8.02 & 0.098 \\
\cline{2-7}
 & \multirow{4}{*}{Contriever} & 8 & 7.37e+12 & 3.0 & 2.26 & 0.185 \\
 &  & 4 & 3.23e+12 & 6.9 & 3.81 & 0.194 \\
 &  & 2 & 1.23e+12 & 18.0 & 5.04 & 0.172 \\
 &  & 1 & 2.54e+11 & 87.5 & 7.32 & 0.125 \\
\cline{2-7}
 & \multirow{4}{*}{TF-IDF} & 8 & 7.35e+12 & 3.0 & 2.32 & 0.188 \\
 &  & 4 & 3.21e+12 & 6.9 & 4.00 & 0.203 \\
 &  & 2 & 1.22e+12 & 18.2 & 5.39 & 0.202 \\
 &  & 1 & 2.38e+11 & 93.4 & 8.01 & 0.159 \\
\cline{2-7}
 & \multirow{4}{*}{Superposition (Ours)} & 8 & 9.66e+11 & 23.0 & 6.52 & 0.271 \\
 &  & 4 & 5.49e+11 & 40.5 & 6.58 & \textbf{0.301} \\
 &  & 2 & 3.42e+11 & 65.0 & 6.59 & 0.288 \\
 &  & 1 & 2.38e+11 & 93.4 & 6.61 & 0.253 \\
\cline{1-7} \cline{2-7}
\multirow{16}{*}{mpt-7b-instruct} & \multirow{4}{*}{BM-25} & 8 & 7.11e+12 & 3.0 & 2.29 & 0.268 \\
 &  & 4 & 3.11e+12 & 7.0 & 4.01 & 0.278 \\
 &  & 2 & 1.18e+12 & 18.4 & 5.38 & 0.269 \\
 &  & 1 & 2.31e+11 & 94.0 & 8.12 & 0.240 \\
\cline{2-7}
 & \multirow{4}{*}{Contriever} & 8 & 7.13e+12 & 3.0 & 2.22 & 0.318 \\
 &  & 4 & 3.13e+12 & 6.9 & 3.80 & 0.334 \\
 &  & 2 & 1.20e+12 & 18.1 & 5.00 & 0.338 \\
 &  & 1 & 2.47e+11 & 87.8 & 7.28 & 0.287 \\
\cline{2-7}
 & \multirow{4}{*}{TF-IDF} & 8 & 7.11e+12 & 3.0 & 2.28 & 0.297 \\
 &  & 4 & 3.11e+12 & 7.0 & 3.99 & 0.332 \\
 &  & 2 & 1.18e+12 & 18.4 & 5.34 & 0.333 \\
 &  & 1 & 2.31e+11 & 94.0 & 8.07 & 0.308 \\
\cline{2-7}
 & \multirow{4}{*}{Superposition (Ours)} & 8 & 9.27e+11 & 23.4 & 5.78 & 0.423 \\
 &  & 4 & 5.28e+11 & 41.0 & 6.14 & 0.456 \\
 &  & 2 & 3.30e+11 & 65.7 & 6.32 & \textbf{0.471} \\
 &  & 1 & 2.31e+11 & 94.0 & 6.46 & 0.465 \\
\cline{1-7} \cline{2-7}
\bottomrule
\end{tabular}
\end{sc}
\end{small}
\end{center}
\end{table}

\subsection{Runtime Optimization Ablation}
\label{appendix:ablations:opt}

In \cref{table:nq_opt_ablation_small} we present a full breakdown of the incremental effect of the path pruning, path caching, and path parallelism optimizations proposed in \cref{section:configurations:pruning}, \cref{section:configurations:caching}, and \cref{section:configurations:parallelization}, respectively.

\begin{table}[H]
\caption{
Ablation of speedup vs. accuracy on the NaturalQuestions-Open dataset by enabling/disabling the optimizations proposed in \cref{section:configurations}.
Each of Pruning?/Caching?/Parallelism? correspond to path pruning, path caching, and path parallelism enabled.
}

\label{table:nq_opt_ablation_small}
\vskip 0.15in
\begin{center}
\begin{small}
\begin{sc}\begin{tabular}{llllccc}

\toprule
 &  &  &  & Compute Cycles & Theor. Speedup & Accuracy \\
Model & Pruning? & Caching? & Parallelism? &  &  &  \\
\midrule
\multirow{8}{*}{OpenELM-3B-In.} & \multirow{4}{*}{False} & \multirow{2}{*}{False} & False & 9.81e+12 & 1.1 & 0.005 \\
 &  &  & True & 6.82e+11 & 15.1 & 0.005 \\
\cline{3-7}
 &  & \multirow{2}{*}{True} & False & 9.86e+11 & 10.5 & 0.005 \\
 &  &  & True & 1.18e+11 & 87.2 & 0.005 \\
\cline{2-7} \cline{3-7}
 & \multirow{4}{*}{True} & \multirow{2}{*}{False} & False & 9.80e+12 & 1.1 & 0.241 \\
 &  &  & True & 6.71e+11 & 15.4 & 0.241 \\
\cline{3-7}
 &  & \multirow{2}{*}{True} & False & 9.74e+11 & 10.6 & 0.241 \\
 &  &  & True & 1.07e+11 & 96.9 & 0.241 \\
\cline{1-7} \cline{2-7} \cline{3-7}
\multirow{8}{*}{bloomz-3b} & \multirow{4}{*}{False} & \multirow{2}{*}{False} & False & 9.08e+12 & 1.1 & 0.114 \\
 &  &  & True & 6.40e+11 & 15.2 & 0.114 \\
\cline{3-7}
 &  & \multirow{2}{*}{True} & False & 9.25e+11 & 10.5 & 0.114 \\
 &  &  & True & 1.18e+11 & 82.4 & 0.114 \\
\cline{2-7} \cline{3-7}
 & \multirow{4}{*}{True} & \multirow{2}{*}{False} & False & 9.06e+12 & 1.1 & 0.223 \\
 &  &  & True & 6.21e+11 & 15.7 & 0.223 \\
\cline{3-7}
 &  & \multirow{2}{*}{True} & False & 9.06e+11 & 10.8 & 0.223 \\
 &  &  & True & 9.92e+10 & 98.2 & 0.223 \\
\cline{1-7} \cline{2-7} \cline{3-7}
\multirow{8}{*}{bloomz-7b1} & \multirow{4}{*}{False} & \multirow{2}{*}{False} & False & 2.18e+13 & 1.0 & 0.125 \\
 &  &  & True & 1.52e+12 & 14.6 & 0.125 \\
\cline{3-7}
 &  & \multirow{2}{*}{True} & False & 2.20e+12 & 10.1 & 0.125 \\
 &  &  & True & 2.68e+11 & 82.8 & 0.125 \\
\cline{2-7} \cline{3-7}
 & \multirow{4}{*}{True} & \multirow{2}{*}{False} & False & 2.18e+13 & 1.0 & 0.253 \\
 &  &  & True & 1.49e+12 & 14.9 & 0.253 \\
\cline{3-7}
 &  & \multirow{2}{*}{True} & False & 2.17e+12 & 10.2 & 0.253 \\
 &  &  & True & 2.38e+11 & 93.4 & 0.253 \\
\cline{1-7} \cline{2-7} \cline{3-7}
\multirow{8}{*}{mpt-7b-instruct} & \multirow{4}{*}{False} & \multirow{2}{*}{False} & False & 2.13e+13 & 1.0 & 0.287 \\
 &  &  & True & 1.46e+12 & 14.8 & 0.287 \\
\cline{3-7}
 &  & \multirow{2}{*}{True} & False & 2.11e+12 & 10.3 & 0.287 \\
 &  &  & True & 2.34e+11 & 92.7 & 0.287 \\
\cline{2-7} \cline{3-7}
 & \multirow{4}{*}{True} & \multirow{2}{*}{False} & False & 2.13e+13 & 1.0 & 0.465 \\
 &  &  & True & 1.46e+12 & 14.8 & 0.465 \\
\cline{3-7}
 &  & \multirow{2}{*}{True} & False & 2.11e+12 & 10.3 & 0.465 \\
 &  &  & True & 2.31e+11 & 94.0 & 0.465 \\
\cline{1-7} \cline{2-7} \cline{3-7}
\bottomrule
\end{tabular}
\end{sc}
\end{small}
\end{center}
\end{table}

\section{Supplementary Algorithm Details}

\subsection{Bayesian Path Selection}
\label{appendix:algorithm:bayesian}
Define $\crossEntName : (\mathbb{R}^{* \times \vocabSize}, \sequences^*) \rightarrow \mathbb{R}$ as the language modeling cross-entropy.
\footnote{
More specifically, this is the ``shifted'' cross entropy between a logit tensor and sequence tensor of the same sequence dimension (where we discard element 1 from the input sequence and the last element from the logit tensor).
}
Formally, for our \textit{ForkJoin} prompt topology, we compute the Bayesian path saliency as follows.

As detailed in \cref{appendix:algorithm:full}, during superposition prompting inference, we compute logits for the preamble $\preambleLogit \in \mathbb{R}^{\seqLen{\preamble} \times \vocabSize}$ (where $\vocabSize$ is the vocabulary size), logits for all documents $\docLogit_i  \in \mathbb{R}^{\seqLen{\doc_i} \times \vocabSize}$, and logits for all queries $\queryLogit_i \in \mathbb{R}^{\seqLen{\query} \times \vocabSize}$ for $i \in \docIndices$.
Then, for each $k \in \docIndices$, we can compute:

\begin{subequations}
\label{equation:bayesian_likelihood}
\begin{align}
    \log P(\doc_i \mid \query_i, \preamble) &= \log P(\query_i \mid \doc_i, \preamble) + \log P(\doc_i \mid \query) - C
    \label{equation:bayesian_likelihood:1}
    \\
    &= \frac{\log \crossEntArgs{\docLogit_i}{\doc_i}}{\seqLen{\doc_i}}
    + \frac{\log \crossEntArgs{\preambleLogit}{\preamble}}{\seqLen{\preamble}} - C
    \label{equation:bayesian_likelihood:2}
\end{align}
\end{subequations}

\cref{equation:bayesian_likelihood:1} follows from Bayes Rule (where $C$ is some unspecified constant), while \cref{equation:bayesian_likelihood:2} follows from our definition of language model. The $C$ term is inconsequential for our use, since we eventually softmax before comparing the likelihoods.
This justifies why we duplicate queries in the \textit{ForkJoin} topology|without duplicating the query for each path, we only have access to $P(\query_i \mid \doc_1, \ldots \doc_{\nOfflineDocs}, \preamble)$, not the independent terms $\{P(\query_i \mid \doc_i, \preamble) \mid i \in \docIndices\}$
We choose to normalize the log likelihood terms by their corresponding sequence lengths ($\seqLen{\doc_i}$ and $\seqLen{\query}$) to effectively achieve a ``likelihood density per token,'' which prevents bias against shorter sequence lengths\footnote{
Note that SGPT \cite{SGPT__Muennighoff2022SGPTGS} avoided length bias by truncating sequences to a common length. We choose not to follow this design decision to prevent potential loss of vital information in the input data.
}.

\subsection{Equilibrium Position Assignment}
\label{appendix:algorithm:position_assignment}

\cref{section:position_assignment} outlined the intuition behind the equilibrium position assignment algorithm.
Here, we algorithmically describe the method.

For convenience, we define the \textbf{ArangePositions} function, $a_\Delta$, as
\begin{equation}
    \label{equation:arange}
    a_\Delta(s_0, m) = \langle s_1, s_1 + \Delta, ..., s_1 + (n - 1) \Delta \rangle \in \posIds^m
\end{equation}
where $\Delta \in \mathbb{R}$ is the step size, $s_0 \in \mathbb{R}$ (the ``starting'' position) and $m \in \mathbb{Z}^+$ (the extend amount).

\begin{algorithm}[ht]
    \caption{Equilibrium Positioning of the ``\textit{Fork}'' portion of the \textit{ForkJoin} Topology}
    \label{alg:equilibrium}
    \begin{algorithmic}
        \STATE {\bfseries Input:} Preamble token sequence $\preamble$, set of document sequences $\docs = \{\doc_i \mid i \in \docIndices\}$
        \STATE $\preamblePos = a_1(0, \seqLen{\preamble})$ \codeComment{Using \cref{equation:arange}}
        \FOR{$i=1$ {\bfseries to} $\nOfflineDocs$}
            \STATE $s_i = S(\docs) / \seqLen{\doc_i}$ \codeComment{Using \cref{equation:distance}}
            \STATE $\docPos_i = a_{s_i}(\max(\preamblePos) + 1, \seqLen{\doc_i})$
        \ENDFOR
        \STATE {\bfseries Output:} Preamble token positions $\preamblePos$, set of document token positions $\{\docPos_i \mid i \in \docIndices\}$
    \end{algorithmic}
\end{algorithm}

\subsection{Full Algorithm Outline}
\label{appendix:algorithm:full}

We denote $\docIndices = \{1, \ldots, \nOfflineDocs\}$ as our document ``indices.''
Note that we stylize KV cache variables as boxed variables for visual distinctness (for instance $\kvCache{a}$).
We define $\emptyKv$ as an ``empty'' KV cache (i.e. sequence length of $0$). We also define $\llmCall : (\kvCache{x}, \inputIds{y}, \posId{y}) \mapsto (\kvCache{y}, \logitVec{\psi})$ where:
\begin{itemize}
    \item $\kvCache{x} \in \kvCaches$ (KV cache used as input)
    \item $\inputIds{y} \in \sequences$ (new tokens not included in KV cache)
    \item $\kvCache{y} \in \kvCaches$ and $\seqLen{\kvCache{y}} = \seqLen{\inputIds{y}}$ (KV cache computed by LLM)
    \item $\logitVec{\psi} \in \mathbb{R}^{\seqLen{\inputIds{y}} \times \vocabSize}$ (logit predictions computed by LLM)
\end{itemize}

Following the insight of \cref{section:configurations:parallelization}, we also define
\begin{equation*}
\begin{split}
\llmCallParallel :\ & (\{\kvCache{x}_i \mid i \in X\}, \inputIds{y}, \posId{y}) 
\mapsto (\{\kvCache{y}_i \mid i \in X\}, \{\logitVec{\psi}_i \mid i \in X\})
\end{split}
\end{equation*}
with analogous outputs to \llmCall, although ``batched'' (note how the call accepts a collection of input KV caches, and outputs a collection of KV caches and a collection of logits).
A visual depiction can be seen in \cref{figure:visual_overview:parallel}.

With these definitions, we present the full formalized preprocessing algorithm \cref{alg:preprocessing} and online serving \cref{alg:serving}.

\begin{algorithm}[ht]
    \caption{Offline Preprocessing}
    \label{alg:preprocessing}
    \begin{algorithmic}
        \STATE {\bfseries Input:} Preamble token sequence $\preamble$
        \STATE {\bfseries Input:} Set of document sequences $\docs = \{\doc_i \mid i \in \docIndices\}$
        \STATE $\preamblePos \oplus \bigoplus_{i \in \docIndices} \docPos_i $ := $\computePosArgs{\preamble, \docs}$ \codeComment{\cref{alg:equilibrium}}
        \STATE $(\preambleKv, \preambleLogit)$ := $\llmCallArgs{\emptyKv}{\preamble}{\preamblePos}$
        \FOR{$i=1$ {\bfseries to} $\nOfflineDocs$}
        \STATE ($\docKv_i, \docLogit_i$) := $\llmCallArgs{\preambleKv}{\doc_i}{\docPos_i}$
        \ENDFOR
        \STATE {\bfseries Output:} Preamble positions $\preamblePos$, KV cache $\preambleKv$, logits $\preambleLogit$
        \STATE {\bfseries Output:} Set of document KV positions $\{\docPos_i \mid i \in \docIndices\}$, KV caches $\{\docKv_i \mid i \in \docIndices\}$ and logits $\{\docLogit_i \mid i \in \docIndices\}$
    \end{algorithmic}
\end{algorithm}

\begin{algorithm}[!ht]
    \caption{Online Serving}
    \begin{algorithmic}
        \label{alg:serving}
        \STATE {\bfseries Input:} Preamble positions $\preamblePos$, KV cache $\preambleKv$, logits $\preambleLogit$
        \STATE {\bfseries Input:} Set of document KV positions $\{\docPos_i \mid i \in \docIndices\}$, KV caches $\{\docKv_i \mid i \in \docIndices\}$ and logits $\{\docLogit_i \mid i \in \docIndices\}$
        \STATE {\bfseries Input:} Query tokens $\query$
        \STATE {\bfseries Input:} Postamble tokens $\task$ \codeComment{For instance, ``\texttt{\textbackslash n\#\#\# Response\textbackslash n}'' if using Alpaca instruct tuning format.}
        \STATE $\queryPos := a_{1}(\max(\docPos_1), \seqLen{\query})$ \codeComment{Using \cref{equation:arange}.}
        \STATE $\inputIds{B}$ := $\{\preambleKv \oplus \docKv_i | i \in \docIndices\}$
        \STATE $\{(\queryKv_i, \queryLogit_i) | i \in \docIndices\}$ := $\llmCallParallelArgs{\inputIds{B}}{\query}{\queryPos}$
        \STATE \codeComment{Could use tuned threshold instead of top-k.}
        \STATE $K := \operatorname*{argmin}^K_{i \in \docIndices} P(\doc_i \mid \query_i, \preamble)$ \codeComment{Using \cref{equation:bayesian_likelihood}.}
        \STATE $\kvCache{\mathit{pdqt}}$ := $\preambleKv \oplus \bigoplus_{k \in K} (\docKv_a \oplus \queryKv_a) \oplus \taskKv$
        \STATE $\taskKv, \taskLogit$ := $\llmCallArgs{\kvCache{\mathit{pdqt}}}{\task}{\taskPos}$
        
        \STATE $\newLogit := \taskLogit$
        
        \STATE \codeComment{Greedy Decoding shown here, but WLOG another decoding scheme could be used.}
        \REPEAT
            \STATE $\new := \textbf{Sample}(\newLogit) \in \sequences$ \codeComment{Some arbitrary sampling procedure.}
            \STATE $(\newKv, \newLogit) := \llmCallArgs{\kvCache{\mathit{pdqt}} \oplus \responseKv}{\new}{\newPos}$
            \STATE $\responseKv := \responseKv \oplus \newKv$
            \STATE $\response := \response \oplus \new$
            \STATE $e := \textbf{Stop?}(\response) \in \{0, 1\}$  \codeComment{Say, 1 if EOS-terminated.}
        \UNTIL $e = 1$
        \STATE {\bfseries Output:} Response token sequence $\response$
    \end{algorithmic}
\end{algorithm}

\section{CUDA Benchmarks}
\label{appendix:cuda}
\begin{table*}[h]
\caption{CUDA speedup measurements for remaining baselines methods not enumerated in \cref{table:nq_topk_ablation} (this table is meant to be directly comparable to \cref{table:nq_topk_ablation}).}
\label{table:nq_cuda}
\vskip 0.15in
\begin{center}
\begin{small}
\begin{sc}\begin{tabular}{llcc}
\toprule
 &  & CUDA Speedup & Accuracy \\
Model & Approach &  &  \\
\midrule
\multirow{3}{*}{bloomz-3b} & Naive LLM-RAG & 1.000 & 0.005 \\
 & Attention Sort & 0.337 & 0.004 \\
 & Prompt Cache & 6.317 & 0.113 \\
\cline{1-4}
\multirow{3}{*}{bloomz-7b1} & Naive LLM-RAG & 1.000 & 0.022 \\
 & Attention Sort & 0.335 & 0.022 \\
 & Prompt Cache & 8.643 & 0.136 \\
\cline{1-4}
\multirow{3}{*}{mpt-7b-instruct} & Naive LLM-RAG & 1.000 & 0.026 \\
 & Attention Sort & 0.332 & 0.028 \\
 & Prompt Cache & 8.113 & 0.278 \\
\cline{1-4}
\bottomrule
\end{tabular}
\end{sc}
\end{small}
\end{center}
\end{table*}

In \cref{table:nq_topk_ablation} and \cref{table:nq_cuda}, we present measurements of the compared methods in a realistic server deployment scenario (an NVIDIA A100 80GB).
Our CUDA implementation is written in pure PyTorch, and we report the median timing over 30 trials for each method (generating a 5 token response to the prompt).
Mirroring our theoretical speedup projections, we report speedups over the Naive LLM-RAG method.

We notice that the actual measured speedups are an order of magnitude smaller than the theoretical maximum speedups calculated.
This is expected|as we heavily optimize the FLOPs (up to 100$\times$), the memory bottlenecks begin to dominate the runtime.
We expect that a fused CUDA kernel implementation could bridge the order of magnitude gap, similar to how \citealp{FlashAttn__Dao2022FlashAttentionFA} achieves an order of magnitude improvement over the naive PyTorch implementation by mitigating memory transfer bottlenecks.

\section{Additional Visualizations}

\subsection{Iterative Superposition}
\label{appendix:algorithm:iterative}
We present a visual depiction of iterative superposition (from \cref{section:experiments:musique}) in \cref{fig:iterative_superposition}.

\begin{figure}[!h]
    \centering
    \includegraphics[width=0.9\linewidth]{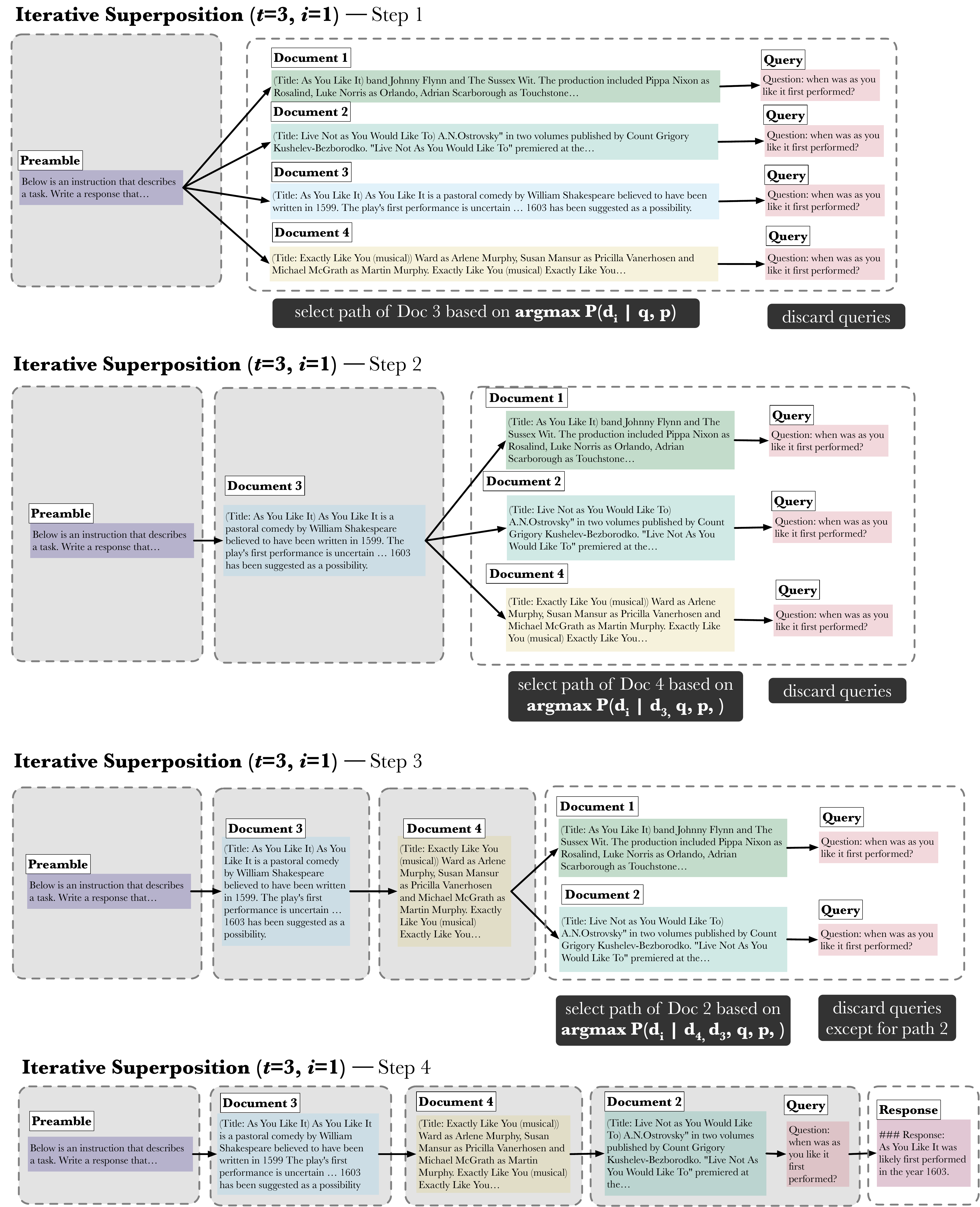}
    \caption{Iterative superposition illustrated example with 3 iterations. Steps 1-3 are the ``superposition'' steps, where we evaluate multiple documents. Step 4 is where we generate the response from the selected (\textit{i.e.} unpruned) paths.}
    \label{fig:iterative_superposition}
\end{figure}

\subsection{Superposition Factor}
We present a visual depiction of superposition factor (from \cref{appendix:ablation:superposition_factor}) in \cref{fig:superposition_quotient}.

\begin{figure}[htb]
    \centering
    \includegraphics[width=\textwidth]{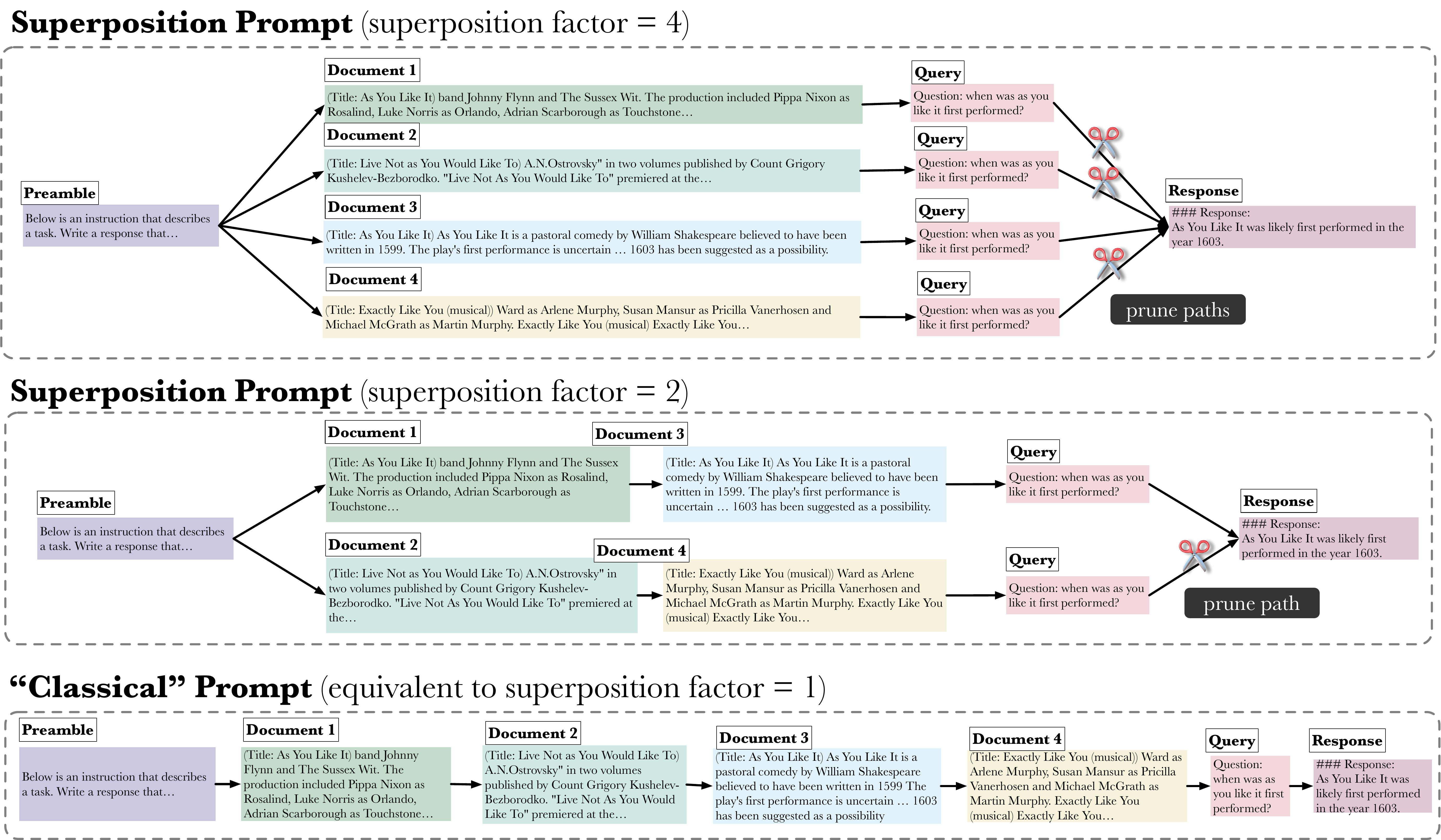}
    \caption{
    Visual depiction of the superposition factor, which is measure of how ``superpositioned'' a prompt is.
    A rigorous definition is provided in \cref{appendix:ablation:superposition_factor}.
    }
    \label{fig:superposition_quotient}
\end{figure}

\section{Dataset Statistics}
\label{appendix:stats}

To get a better sense of the effect of the position encoding schemes, we present token count statistics for the document lengths of each evaluation dataset (using a BPE-based tokenizer).
Define $\nOfflineDocs$ to be the number of offline documents per example, and define $M$ to be the overall number of examples.
Use $\doc_i^j$ to correspond to the $i$th document within the $j$th example.

\begin{itemize}
    \item Average document length ($\frac{1}{M \nOfflineDocs} \sum_i \sum_j \seqLen{\doc_i^j}$)
    \begin{itemize}
        \item Natural Questions: 142.6
        \item MuSiQue: 121.2
    \end{itemize}
    \item Average length of longest document ($\frac{1}{M} \sum_j \max_i \seqLen{\doc_i^j}$)
    \begin{itemize}
        \item Natural Questions: 170.2
        \item MuSiQue: 222.8
    \end{itemize}
    \item Average Left Align token padding gap size ($\frac{1}{M \nOfflineDocs} \sum_j \sum_i (\max_k \seqLen{\doc_k^j} - \seqLen{\doc_i^j})$)
    \begin{itemize}
        \item Natural Questions: 27.6
        \item MuSiQue: 149.6
    \end{itemize}
\end{itemize}

\section{Prompt Example on NaturalQuestions-Open}
\label{appendix:prompt_examples}
As LLMs are sensitive to the specific wording of prompts, we present a prompt example in full for reproducibility.

\begin{figure}[ht]
\colorbox{lightgray}{
\begin{minipage}{\textwidth}
Below is an instruction that describes a task. Write a response that appropriately completes the request.
\newline
\newline
\#\#\# Instruction:
\newline
Write a high-quality answer for the given question using only the following relevant search results.
\newline
\newline
[Document](Title: The Crossing (play)) The Crossing (play) The Crossing is a 2006 South African one-man play by Jonathan Nkala.[\ldots]
\newline
[Document](Title: The Crossing (TV series)) The Crossing is an American science fiction thriller series that airs on ABC and CTV. The series debuted on April 2, 2018. On March 20, 2018, ABC released the pilot episode on their website. The series is filmed in British Columbia, Canada.
\newline
[Document](Title: Crossing South) Crossing South Crossing South is a travel show, television production that was created[\ldots]
\newline
[Document](Title: Crossing (2007 film)) He received a Gemini nomination for his work on the show. Crossing (2007 film) Crossing is a 2007[\ldots]
\newline
[Document](Title: The Crossing (TV series)) British Columbia and in New Westminster. The first camp footage was filmed at Camp McLean. Filming in Vancouver[\ldots]
\newline
[Document](Title: Crossing East) honored by winning a Peabody Award. Crossing East Crossing East is an American documentary series for public radio produced by Dmae Roberts[\ldots]
\newline
[Document](Title: Crossings (TV series)) Crossings (TV series) Crossings is a Malaysia dark comedy television drama that consisted of 13 episodes. Bob works as a copywriter[\ldots]
\newline
[Document](Title: The Mexican Dream) of the border crossing action takes place was not that difficult. The bar and carwash were probably[\ldots]
\newline
[Document](Title: Southern Crossing (film)) it was filmed was "so wonderful" they had to demolish it in references to the theater's heritage factor.[\ldots]
\newline
[Document](Title: Crossing East) Crossing East Crossing East is an American documentary series for public radio produced by Dmae Roberts and MediaRites and hosted by George Takei and Margaret Cho. Covering Asian immigration to the[\ldots]
\newline
[Document](Title: Crossing Lines) series, having previously produced miniseries, as well as its first project since being acquired by StudioCanal in 2012.[\ldots]
\newline
[Document](Title: The Crossing (TV series)) a threat. Set in the fictional town of Port Canaan, Oregon and in Seattle, the series was filmed in coastal areas of British Columbia[\ldots]
\newline
.
\newline
.
\newline
.
\newline
[Document](Title: The Crossing Hero) Fridays, at 12nn. Beginning 5 April 2015, "The Crossing Hero" airs on Taiwan's China Television (CTV),[\ldots]
\newline
\newline
Question: where did they film the show the crossing?
\newline
\newline
\#\#\# Response:
\newline
$<$model continues from here...$>$
\end{minipage}
}
\end{figure}

\newcommand{\dottt}{[\ldots]}

\section{Prompt Example on MuSiQue}

\begin{figure}[ht]
\colorbox{lightgray}{
\begin{minipage}{\textwidth}
Below is an instruction that describes a task. Write a response that appropriately completes the request.
\newline
\newline
\#\#\# Instruction:
\newline
You are a question-answering assistant, who is careful to reference source material. Use the source(s) below to answer the user question.
\newline
\newline
[Document](Title: National Workers Memorial (Australia)) The National Workers Memorial in the national capital, Canberra, Australian Capital\dottt
\newline
[Document](Title: Braddon, Australian Capital Territory) Braddon (postcode: 2612) is an inner north suburb of Canberra, Australian Capital\dottt
\newline
[Document](Title: WKDM) WKDM 1380 is a United States ethnic brokered radio station licensed to New York City. The station is owned by Multicultural Broadcasting and airs programming in Mandarin Chinese, 24 hours a day from Monday\dottt
\newline
[Document](Title: York, Upper Canada) The Town of York was the second capital of the district of Upper Canada and the predecessor to Toronto (1834). It was established in 1793 by Lieutenant - Governor John Graves Simcoe as a ``temporary'' location for the capital of Upper Canada, while he made plans to build a capital near today's\dottt
\newline
[Document](Title: KLIF-FM) KLIF-FM (93.3 FM, branded as ``Hot 93.3'') is a radio station licensed to serve Haltom City, Texas, United States. The station is owned by Cumulus Media, and the broadcast license is held by Radio License\dottt
\newline
[Document](Title: WRGV) WRGV (107.3 FM) is a radio station licensed to serve the community of Pensacola, Florida, United States. The station is currently owned by iHeartMedia, Inc. and the broadcast license is held by Clear Channel Broadcasting Licenses, Inc. WRGV broadcasts an urban contemporary music format to the greater\dottt
\newline
[Document](Title: WWRU) WWRU is a Korean language AM radio station licensed to Jersey City, New Jersey, broadcasting to the New York\dottt
\newline
[Document](Title: KDBS) KDBS (1410 AM, ESPN Alexandria) is an American radio station broadcasting a sports talk format. The station is licensed by the Federal Communications Commission (FCC) to serve the community of Alexandria, Louisiana. The\dottt
\newline
[Document](Title: Brantley York) Richard Brantley York (January 3, 1805 – October 7, 1891) was a Methodist minister and educator best known for founding and serving as president of the institution that would become Duke\dottt
\newline
.
\newline
.
\newline
.
\newline
[Document](Title: Randolph County, Illinois) Owing to its role in the state's history, the county motto is "Where Illinois Began." It contains the historically\dottt
\newline
[Document](Title: Minsk Region) Minsk Region or Minsk Voblasć or Minsk Oblast (, "Minskaja vobłasć" ;, "Minskaja oblastj") is one of the regions of Belarus. Its administrative center is Minsk, although it is a separate administrative\dottt
\newline
[Document](Title: Mount Franklin (Australian Capital Territory)) Mount Franklin is a mountain with an elevation of in the Brindabella Ranges that is located on the border\dottt
\newline
\newline
Question: When did the town WIZE is licensed in become capitol of the state where Brantley York was born?
\newline
\newline
\#\#\# Response:
\newline
$<$model continues from here...$>$
\end{minipage}
}
\end{figure}

\end{document}